\documentclass{article}

    \PassOptionsToPackage{numbers, compress}{natbib}
 \usepackage[preprint]{neurips_2026}

\usepackage[utf8]{inputenc} 
\usepackage[T1]{fontenc}    
\usepackage{hyperref}       
\usepackage{url}            
\usepackage{booktabs}       
\usepackage{amsfonts}       
\usepackage{nicefrac}       
\usepackage{microtype}      
\usepackage{xcolor}         

\usepackage{graphicx}
\usepackage{amsmath} 
\usepackage[capitalize]{cleveref}
\crefname{section}{Sec.}{Secs.}
\Crefname{section}{Section}{Sections}
\Crefname{table}{Table}{Tables}
\crefname{table}{Tab.}{Tabs.}
\crefname{algorithm}{Alg.}{Algs.}
\Crefname{algorithm}{Algorithm}{Algorithms}

\usepackage{algorithm}    
\usepackage{algorithmic}
\usepackage{amssymb}    
\usepackage{booktabs} 
\usepackage{xcolor}
\usepackage{multirow}
\usepackage{graphicx}
\usepackage{tcolorbox}
\usepackage{listings}
\tcbuselibrary{breakable}
\usepackage{tcolorbox}
\usepackage{fvextra}
\tcbuselibrary{breakable}

\newtcolorbox{taskblock}[2][]{%
  center title,
  fonttitle=\bfseries\footnotesize,
  fontupper=\scriptsize,
  boxrule=0.8pt,
  arc=3pt,
  left=1pt,right=1pt,top=1pt,bottom=1pt,
  title=#2,
  halign=center,
  before skip=4pt,
  after skip=4pt,
  colback=gray!5!white,
  colframe=black!50!gray,
  breakable,
  #1
}

\usepackage{algorithm}  
\usepackage{algorithmic}
\usepackage{amssymb}
\usepackage{xcolor}     

\title{WorldSR: Harnessing World Knowledge Search for Grounded Image Super-Resolution}

\author{%
  Junyuan Deng$^1$ \; Xinyi Wu$^1$ \;
  Zhenyao Wu$^1$\thanks{Corresponding authors.} \; Song Wang$^{1,2*}$ \;
  Yu Wang$^1$ \; Congchao Zhu$^1$ \\[1ex]
  $^1$Honor Device Co., Ltd.
  \quad
  $^2$Shenzhen University of Advanced Technology\\[0.5ex]
  \texttt{\{dengjunyuan1, wuxinyi, wuzhenyao, wangyu24, zhucongchao\}@honor.com} \\
  \texttt{wangsong@suat-sz.edu.cn} \qquad \qquad \href{https://gwanjyun.github.io/worldsr.github.io/}{\textbf{Project Page}}
}

\begin{document}

\maketitle
\begin{figure}[h]
    \centering
    \vspace{-25pt}
    \includegraphics[width=1.0\textwidth]{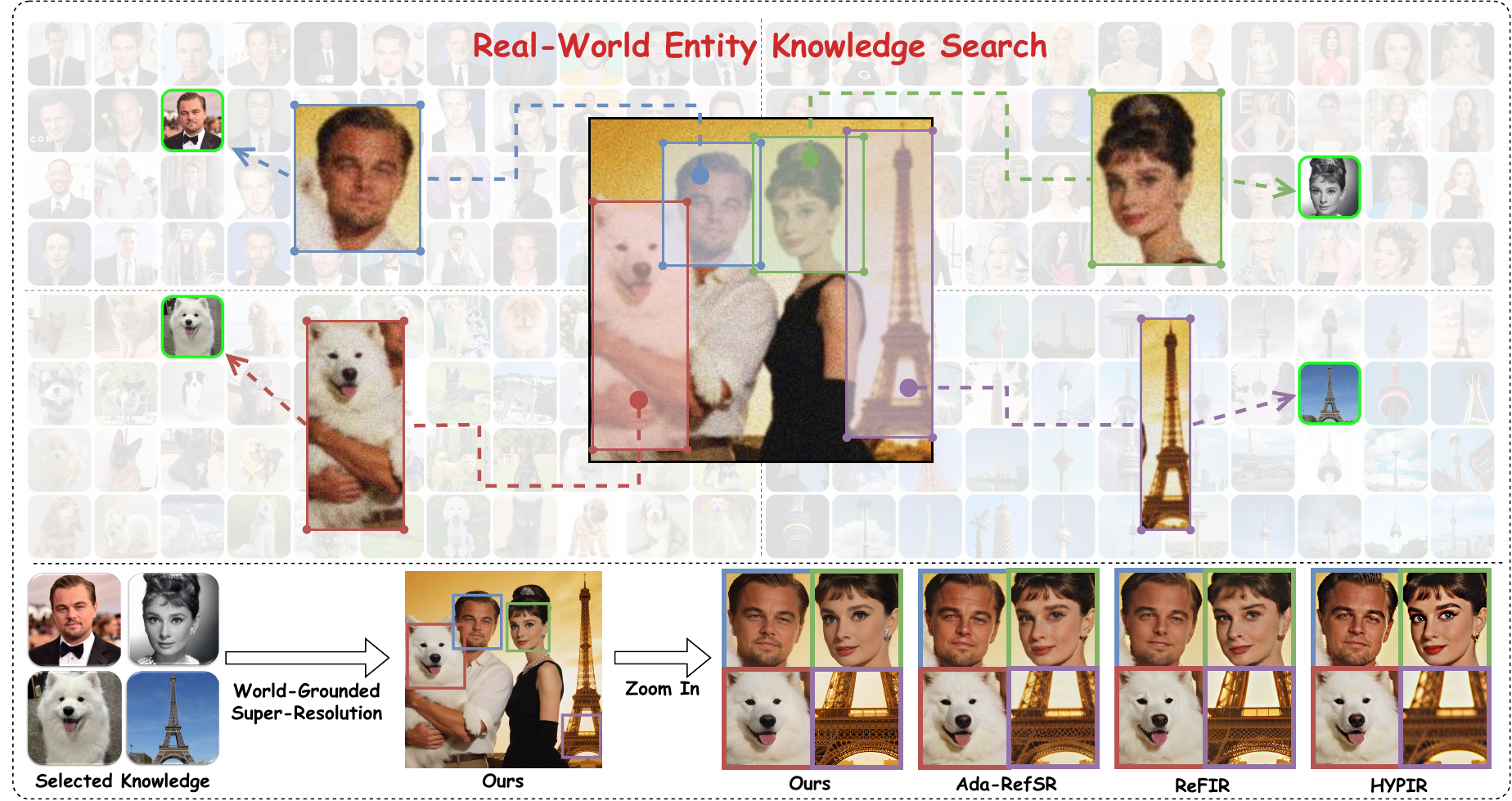}
    \vspace{-15pt}
    \caption{We propose WorldSR, a framework that integrates explicit, entity-aligned visual world knowledge into the super-resolution to mitigate hallucinations and enhance super-resolution quality.
    }
    \label{fig:teaser}
    \vspace{-5pt}
\end{figure}

\begin{abstract}
Generative-prior-based super-resolution (SR) methods leverage implicit knowledge learned from large-scale models to enhance low-resolution inputs. However, such priors are inherently stochastic and optimized for diverse generation, conflicting with the deterministic, high-fidelity reconstruction required for SR. It is also tied to the base model, costly to update, and prone to hallucinations as real-world data evolves. To address these limitations, we introduce WorldSR, a novel framework that augments SR with explicit, entity-aligned visual world knowledge. WorldSR automatically decomposes the input into semantic entities and performs entity-wise retrieval of structurally and semantically aligned images from a large-scale external corpus. For efficiency and reliability, we incorporate a memory mechanism to reuse previously retrieved results and a multi-step filtering process to refine candidate references. The selected references as world knowledge are then integrated through a novel multi-reference super-resolution model, which provides fine-grained constraints for detail synthesis and structural recovery. Experiments on real-world SR benchmarks show that WorldSR consistently improves structural fidelity and visual quality, outperforming existing generative super-resolution methods. Our results suggest that entity-level grounding in external visual world knowledge provides a simple and effective complement to model-bound generative priors.

\end{abstract}
\section{Introduction}\label{sec:Introduction}
Image super-resolution aims to reconstruct high-resolution images from low-resolution inputs by recovering lost fine details.
Recent advances leverage powerful generative priors embedded in large-scale text-to-image models~\cite{rombach2022high, esser2024scaling, labs2025flux1kontextflowmatching, wu2025qwen, cai2025z}, which are trained on massive datasets and encode rich general visual knowledge in their parameters. 
This implicit knowledge enables recent works~\cite{wang2024exploiting, wu2024seesr, yu2024scaling, ai2024dreamclear, wu2024one, sun2025pixel, lin2025harnessing, wu2025omgsr} to enrich low-resolution inputs with visually plausible details, substantially boosting perceptual quality.
However, these globally applied generative priors lack explicit grounding in real-world semantics, often leading to hallucinated details~\cite{ren2025hallucination, kim2025hallugen, wang2022faceformer} that are visually plausible but semantically incorrect.

To mitigate hallucinations, recent approaches explore multimodal large language models (MLLMs) to provide textual descriptions of low-resolution inputs~\cite{sun2024coser, huang2024instantir, lee2025reference}. 
These descriptions are then used to synthesize semantically aligned images that serve as references to guide the diffusion-based super-resolution process, improving semantic consistency to some extent. Nevertheless, such caption-derived references are prone to out-of-distribution artifacts, which can introduce structural inconsistencies and degrade the fidelity of the final super-resolved outputs.

More direct approaches~\cite{guo2024refir, yan2025rasr, shi2025ultra, wang2026trust} leverage reference images with high visual similarity (\emph{e.g.}, different views of the same subject or scene) to guide the super-resolution process. In domain-specific settings such as face restoration~\cite{zhang2025instantrestore, hsiao2024ref, varanka2024pfstorer, jois2024reference}, multiple reference images are sometimes employed to aggregate complementary facial details.
However, these methods typically rely on a single or a small set of pre-collected references, significantly restricting their ability to exploit diverse and complementary visual cues. Additionally, in complex scenes with multiple semantic entities, the lack of region-level correspondence between inputs and references often leads to structural inconsistencies and hallucinated details, thereby limiting their robustness in real-world scenarios.

In recent years, retrieval-augmented approaches have shown strong effectiveness in mitigating hallucinations by grounding model predictions in external, dynamically retrieved knowledge, thereby improving factual consistency in MLLM-based vision-language tasks~\cite{lewis2020retrieval, qu2025alleviating, fang2025guided, sarwar2025filterrag}.  
However, directly extending this paradigm to image super-resolution is non-trivial, as existing methods typically rely on holistic image-level similarity, which is insufficient for handling complex scenes with multiple semantic entities and fine-grained region-level variations.
To address this limitation, we propose \textbf{WorldSR}, an MLLM-empowered agent-style framework that reformulates image super-resolution as an entity-level grounding problem using explicit visual world knowledge.

Specifically, WorldSR first leverages an MLLM to parse the input low-resolution image and identify real-world entities, enabling structured semantic understanding. It then retrieves up-to-date reference images from web sources and local databases conditioned on these entities, and further filters them based on entity-level visual similarity and quality. Finally, a world-grounded super-resolution model is carefully designed to integrate the selected multi-reference knowledge into the reconstruction process. To improve efficiency, a memory verifier is introduced to cache previously retrieved results and avoid redundant queries. Through iterative refinement, WorldSR progressively reduces hallucinations and produces outputs that are better aligned with real-world semantics. We evaluate WorldSR on multiple benchmark datasets and demonstrate consistent improvements over state-of-the-art super-resolution methods in visual quality, structural fidelity, and semantic consistency. Both quantitative and qualitative results show that WorldSR is effective in reconstructing well-defined objects and real-world landmarks in complex scenes, while significantly reducing hallucinated content.

\section{Related Work}\label{sec:Related Work}
\subsection{Prior-Based Image Super-Resolution}
Conventional image SR methods~\cite{dong2015image, ledig2017photo, wang2018esrgan, wang2021real, liang2021swinir} typically rely on local structural priors and realistic degradation simulation pipelines~\cite{wang2021real} to tackle the challenges associated with real-world super-resolution. However, these methods often face significant difficulties in generating high-frequency details that faithfully reproduce natural textures. 
StableSR~\cite{wang2024exploiting} was the first to incorporate generative priors by adapting a pretrained text-to-image (T2I) model for super-resolution, substantially improving the high-frequency content of SR outputs. 
Following this, generative priors from pretrained T2I models have been widely integrated into contemporary SR architectures~\cite{wu2024seesr, yu2024scaling, ai2024dreamclear, wu2024one, sun2025pixel, lin2025harnessing, wu2025omgsr}, marking a significant advancement in SR research. 
Nevertheless, with these generative capabilities, the issue of hallucinations in SR results has become increasingly prominent. 
To mitigate this challenge, CoSeR~\cite{sun2024coser} proposes guiding SR models with semantically aligned reference images. InstantIR~\cite{huang2024instantir} and iRAG~\cite{lee2025reference} generate reference images semantically similar to the input image using T2I models, which are subsequently used to constrain and guide the super-resolution process. Recently, there has been a growing trend to shift from semantic priors toward texture priors. ReFIR~\cite{guo2024refir} introduces a plug-and-play, training-free approach to inject reference image textures into the restoration pipeline, while RASR~\cite{yan2025rasr} leverages matched references as spatial conditions for ControlNet, facilitating high-fidelity texture transfer. TriFlowSR~\cite{shi2025ultra} designs a dual-branch architecture for landmark SR, jointly encoding low-resolution content and high-resolution reference textures. And Ada-RefSR~\cite{wang2026trust} incorporates adaptive correlation gating into attention-based reference fusion, striking a balance between fidelity, naturalness, and efficiency.
However, despite these advances, existing reference-based SR methods remain fundamentally limited by their reliance on single-source, globally shared priors.

\begin{figure}[t]
    \centering
    \includegraphics[width=1.0\textwidth]{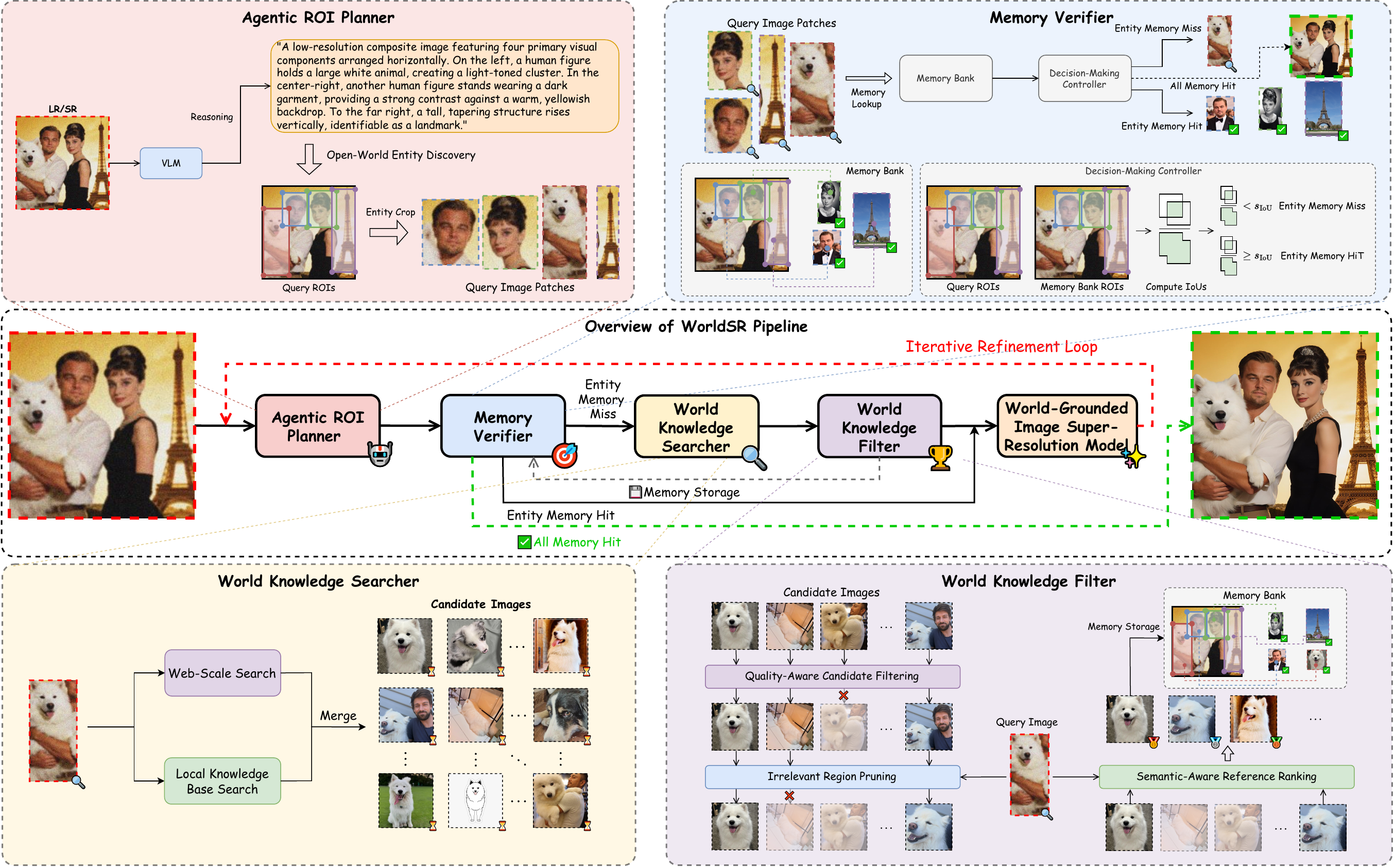}
    \vspace{-20pt}
    \caption{Overview of the WorldSR pipeline.}
    \label{fig:WorldSR_pipeline}
\end{figure}
\subsection{Agentic Systems for Visual Tasks}
Agentic systems have revolutionized visual task processing by introducing autonomous reasoning, tool-use capabilities, and multimodal collaboration to address complex visual challenges. In the realm of high-level visual generation and editing, JarvisArt~\cite{linjarvisart} leverages LLM-powered agents to automatically interpret user instructions and operate Lightroom tools for professional image retouching, while Gen-Searcher~\cite{feng2026gen} equips agents with internet search capabilities to autonomously gather up-to-date text-image pairs for refreshing training datasets. Additionally, Unify-Agent~\cite{chen2026unify} proposes an agentic image generation pipeline of five steps: world-grounded image synthesis, encompassing prompt understanding, multimodal evidence searching, grounded recaptioning, and final synthesis. 
Regarding low-level vision tasks, RestoreAgent~\cite{chen2024restoreagent} leverages an agent to autonomously assess the type and extent of degradation in input images and then executes a restoration pipeline using multiple restoration models. Similarly, AgentIR~\cite{zhuintelligent} follows this design paradigm, but further introduces a rollback mechanism to determine the optimal restoration strategy. JarvisIR~\cite{lin2025jarvisir} constructs a large-scale instruction-response dataset to fine-tune the underlying MLLM, thereby improving overall agentic system robustness. Furthermore, MAIR~\cite{jiang2026multi} improves restoration performance by scaling up the number of agents to optimize pipeline scheduling, and 4KAgent~\cite{zuo4kagent} expands its restoration model zoo to further enhance the agent's capability to upscale images to 4K resolution. 
All the aforementioned works~\cite{chen2024restoreagent, zhuintelligent, lin2025jarvisir, jiang2026multi, zuo4kagent} utilize agents to invoke specific neural networks for targeted image restoration. In contrast, this paper presents an agent-style framework to acquire world knowledge and mitigate generative hallucination issues in super-resolution.

\section{WorldSR}\label{sec:AgentRefSR}
\subsection{Overview}
We introduce WorldSR, a novel agent-style grounded image super-resolution framework that integrates world knowledge search to effectively mitigate hallucinations and improve super-resolution quality. As shown in~\cref{fig:WorldSR_pipeline}, the overall pipeline is an iterative process that undergoes multiple refinements. It consists of five key components, each responsible for a specific task: planning, verification, searching, filtering, and super-resolution. 

The \textbf{Agentic ROI Planner}~(\cref{subsec:planning}) leverages a MLLM to analyze the image content and identify potential regions of interests (ROIs) that connect visual entities to real-world knowledge. The \textbf{Memory Verifier}~(\cref{subsec:verification}) queries the Memory Bank using the extracted ROI knowledge bounding boxes.
If any entity memory is missing, the subsequent search and filter module will be triggered. If all entities are hit, the SR result generated in the previous round is directly returned as the final output.
The \textbf{World Knowledge Searcher}~(\cref{subsec:searching}) retrieves relevant candidate images for all ROI regions from web and local databases. The \textbf{World Knowledge Filter}~(\cref{subsec:filtering}) further selects the optimally matched image from these candidates images via image quality assessment, query-aware similar content extraction, and reference-guided ranking. Ultimately, the \textbf{World-Grounded Image Super-Resolution Model}~(\cref{subsec:SR}) injects all optimally matched real-world knowledge into the low-resolution input to generate a realistic and world-grounded super-resolution output. 

\subsection{Agentic ROI Planner}\label{subsec:planning}
The Agentic ROI Planner module comprises two core processes: entity-level semantic reasoning and the extraction of corresponding ROI bounding boxes. Specifically, we first leverage an MLLM as the core perception tool to perform entity-level semantic reasoning on the input image $I$, enabling a comprehensive understanding of real-world knowledge entities. Subsequently, the MLLM produces a set of bounding boxes $\{B_{i}\}_{i=1}^{N}$, leveraging its inherent capability for open-world entity discovery. These generated bounding boxes are constrained by a carefully designed system prompt to ensure that the regions are precise and their contents are grounded in real-world knowledge. Guided by these bounding boxes, $N$ corresponding image patches $\{P_i\}_{i=1}^{N}$ are automatically cropped from the input image for the downstream Memory Verifier and World Knowledge Searcher. 
However, extracting real-world entities from low-resolution images remains challenging.
To mitigate this, we introduce an iterative refinement loop that leverages the intermediate SR results as visual feedback to progressively suppress hallucinations, as shown in \cref{fig:WorldSR_pipeline}. Formally, the input image $I$ at each step $n$ is either the initial low-resolution image $I^{(0)}$ or the rolled-back intermediate super-resolution output $I^{(n-1)}$ from the preceding iteration. For simplicity, we omit the iteration superscript $(n)$ hereafter.

\subsection{Memory Verifier}\label{subsec:verification}
Since our framework employs an iterative refinement loop strategy to progressively suppress hallucinations, the Agentic ROI Planner may generate overlapped ROIs across successive iterations. To manage these overlapping regions and avoid redundant computations, we propose the Memory Verifier. This module includes two core components: a Memory Bank for retaining historical context and a Decision-Making Controller for adaptive action routing.

\paragraph{Memory Bank} 
The Memory Bank is designed to cache previously explored regions along with their associated contextual knowledge. Specifically, it stores ROI bounding boxes together with their corresponding retrieved and filtered world knowledge, organized as a set of paired entries $(B_j, R_j^*)$.

\paragraph{Decision-Making Controller}  
The Decision-Making Controller determines the appropriate action routing for the current iteration. It begins by calculating the Intersection over Union (IoU) between the newly proposed query ROI $B_{i}$ and all ROIs stored in the Memory Bank. If the maximum IoU score $s_i^*$ exceeds a pre-defined threshold $s_{\text{IoU}}$, the controller bypasses redundant operations and outputs its corresponding knowledge $R_i^*$. Otherwise,  the controller routes the queried image patch $P_i$ to the World Knowledge Searcher to execute a new retrieval operation. Furthermore, if all newly queried ROIs result in memory hits, it indicates that the iterative refinement has converged. Accordingly, the system terminates the loop and outputs the SR image generated in the preceding iteration.

\subsection{World Knowledge Searcher}\label{subsec:searching}
The World Knowledge Searcher is responsible for retrieving external real-world knowledge relevant to the queried image, providing grounded visual evidence to mitigate hallucinations during super-resolution. To ensure both comprehensive and highly accurate information retrieval, we adopt a dual-engine retrieval strategy that integrates web-scale search and local knowledge base search.

\paragraph{Web-scale Search} 
For the image patch $P_i$ output by the Agentic ROI Planner, a web-scraping tool $\mathcal{F}_{\text{web}}$ automatically uploads the image patch $P_i$ to the public internet and utilizes reverse image search technology to retrieve the top-$K_{\text{web}}$ most similar images $\{R_i^{j}\}_{j=1}^{K_{\text{web}}} = \mathcal{F}_{\text{web}}(P_i)$.
In our practical implementation, we use Google Lens as the mainstream image search tool, yielding the top-$K_{\text{local}}$ results (each containing a source URL and a thumbnail). An automated web scraper tool then accesses these URLs to retrieve the associated high-resolution source images.

\paragraph{Local Knowledge Base Search} 
For the pre-registered local knowledge base $\mathcal{D}_{\text{local}}$, we employ a dedicated retrieval algorithm. Each image in the local database is preprocessed by a robust visual encoder (\emph{e.g.}, SigLIP-2~\cite{tschannen2025siglip}) to generate the semantic embedding. Given the query image patch $P_i$, it is first encoded using the same visual encoder. The resulting embedding is then compared to all stored entries in the local database by computing cosine similarity. This allows the retrieval of the top-$K_{\text{local}}$ most similar images, denoted as $\{R_i^{j}\}_{j=1}^{K_{\text{local}}} = \mathcal{F}_{\text{local}}(P_i, \mathcal{D}_{\text{local}})$.

Finally,  we combine the web-based candidates ($\{R_i^{j}\}_{j=1}^{K_{\text{web}}}$) and local candidates ($\{R_i^{j}\}_{j=1}^{K_{\text{local}}}$) into a unified candidate set, denoted as $\{R_i^{j}\}_{j=1}^{K}=\{R_i^{j}\}_{j=1}^{K_{\text{web}}}  \cup \{R_i^{j}\}_{j=1}^{K_{\text{local}}}$.

\subsection{World Knowledge Filter}\label{subsec:filtering}
Since the retrieved images originate from diverse sources and acquisition pipelines, the candidate set often exhibits various forms of image degradation, partially irrelevant content, and insufficient semantic relevance. To address these issues, we introduce a dedicated World Knowledge Filter module to select the optimal reference image $R_i^*$ that best aligns contextually with the query patch $P_i$. Note that $R_i^*$ and its associated ROI bounding box $B_i$ are stored as a pair in the memory bank.

\paragraph{Quality-Aware Candidate Filtering} 
To maintain the high visual quality of the candidate images, we introduce a Quality-Aware Candidate Filtering strategy, which employs dedicated image quality assessment models for reliable and robust quality scoring. Specifically, we utilize RALI~\cite{zhao2025reasoning} and Q-Insight~\cite{li2025qinsight} to compute image quality scores. Any candidate image receiving a score lower than the predefined threshold $s_{\text{filter}}$ from either model is strictly discarded.

\paragraph{Irrelevant Region Pruning}
Some candidate images contain regions unrelated to the query. To remove such noise, we leverage an MLLM to localize query-relevant regions in each candidate image. Specifically, the query image, candidate image, and a grounding prompt are jointly fed into the model to predict bounding boxes that align semantically with the target. The candidate images are then cropped accordingly, and the resulting patches are forwarded to the subsequent module.

\paragraph{Semantic-Aware Reference Ranking}
To further select the world knowledge that best matches the query semantics, we employ a Semantic-Aware Reference Ranking module to evaluate the remaining candidates. Specifically, a multimodal ranking model (\emph{e.g.}, Qwen3-VL-Reranker~\cite{li2026qwen3}) is used to compute semantic similarity scores between the query and the cropped candidates. The top-$1$ candidate, ranked by semantic similarity, is then selected as the optimal reference $R_i^*$.

\subsection{World-Grounded Image Super-Resolution Model}\label{subsec:SR}
\begin{figure}[t]
    \centering
    \includegraphics[width=1.0\textwidth]{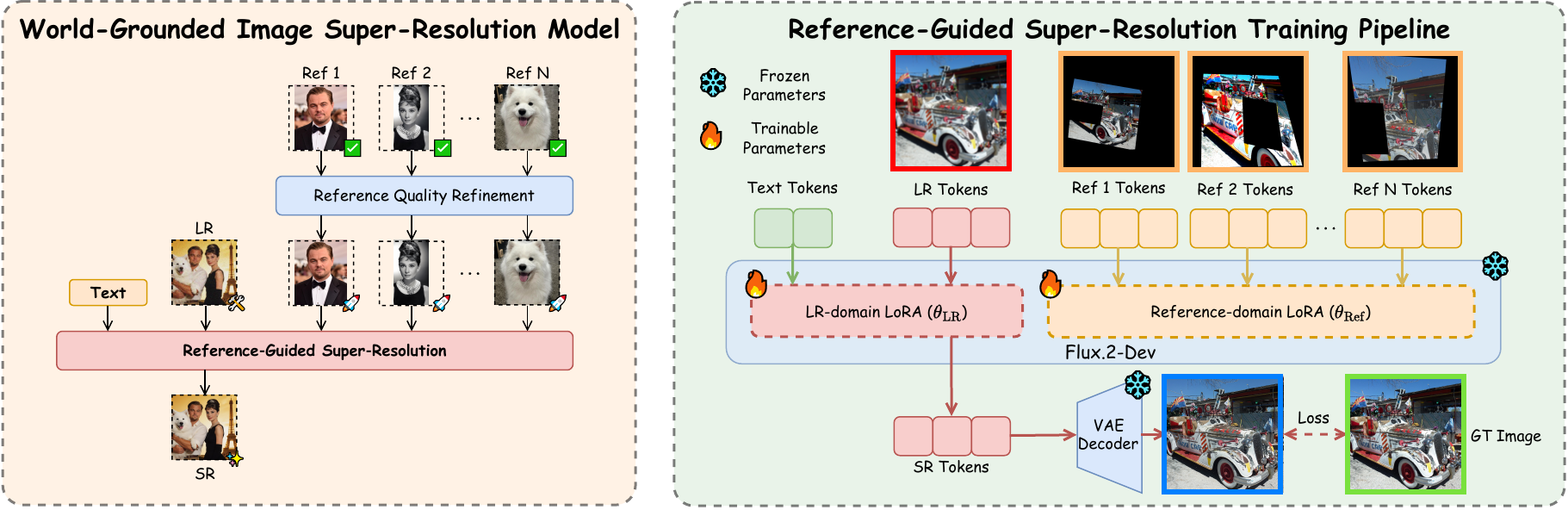}
    \vspace{-15pt}
    \caption{The World-Grounded Image Super-Resolution Model and its training pipeline.}
    \label{fig:WorldSR_train_infer}
    \vspace{-10pt}
\end{figure}

Given a set of world knowledge $\{R_i^*\}_{i=1}^{N}$ aligned with the grounded LR input, we carefully design a novel world-grounded image super-resolution model. As shown in \cref{fig:WorldSR_train_infer}, the model comprises two key components: Reference Quality Refinement and Reference-Guided Super-Resolution. 

\paragraph{Reference Quality Refinement} 
Although the reference images have been filtered by the preceding module, they may still exhibit complex real-world degradations that are not fully captured by IQA metrics. 
To address this issue, we introduce a dedicated Reference Quality Refinement (RQR) module to further enhance the visual quality of the retrieved references.
Specifically, the model follows the paradigm of recent one-step adversarial SR approaches~\cite{wu2024one, sun2025pixel, lin2025harnessing, wu2025omgsr}, and is built upon a pretrained text-to-image DiT backbone~\cite{flux-2-2025}, augmented with a Reference Quality Refinement LoRA ($\theta_{\text{RQR}}$). This design enables efficient restoration while preserving fine-grained high-frequency details, yielding higher-quality and more reliable references for downstream reconstruction.

\paragraph{Reference-Guided Super-Resolution} 
To effectively transfer rich textures from the refined reference images to the target LR image, we formulate the reference-guided super-resolution as unified sequence modeling. Specifically, the LR image $I_{\text{LR}}$ and the refined references $\{\hat{R}_i^*\}_{i=1}^{N}$ are encoded into latent tokens, flattened, and concatenated along the sequence dimension. 
Since the degraded LR tokens and high-quality reference tokens originate from distinct data distributions, sharing identical model parameters restricts the ability of the one-step super-resolution network to effectively disentangle these domains. To bridge this domain gap, we introduce a dual-LoRA architecture (see \cref{fig:WorldSR_train_infer}) to explicitly decouple domain-specific representations. A dedicated LR-domain LoRA ($\theta_{\text{LR}}$) processes LR tokens, while a reference-domain LoRA ($\theta_{\text{Ref}}$) encodes reference tokens.
By projecting the tokens into a unified feature domain, the proposed architecture enables unconstrained global interactions. 
This design allows the model to implicitly learn spatial correspondences and effectively inject high-fidelity world knowledge from multiple reference images into the LR representation, resulting in more faithful texture reconstruction and structure preservation.

\paragraph{Implementation Details}
To efficiently assign the proper LoRA adapters to arbitrary tokens, our proposed dual-LoRA architecture in Reference-Guided Super-Resolution model follows the adapter modulation strategy of OmniControl~\cite{tan2025ominicontrol}. We adopt a one-step training strategy where only LoRA parameters are optimized.
The training objective combines pixel-level and perceptual supervision, including the MSE loss $\mathcal{L}_{2}$, the OC-EA-DISTS perceptual loss~\cite{wu2025omgsr} $\mathcal{L}_{\text{p}}$, and a DINOv3-based~\cite{simeoni2025dinov3} adversarial loss $\mathcal{L}_{\text{g}}$. 
To improve robustness under varying reference conditions, we randomly sample the number of reference images from $0$ to $N_{\text{max}}$ during training. In the zero-reference case, all reference inputs are dropped, and only the LR-domain LoRA parameters $\theta_{\text{LR}}$ are updated.
More implementation details are provided in \cref{subsec:more_implementation_details} of the supplementary material.

\section{Experiments}\label{sec:Experiments}
\subsection{Experimental Settings}
Across all our experiments, we adopt Qwen3.5-35B-A3B~\cite{qwen3.5} as the unified multimodal vision-language model, and leverage the state-of-the-art text-image-to-image model, Flux.2-dev~\cite{flux-2-2025}, to serve as a generative prior for our world-grounded image super-resolution model. The QLoRA~\cite{dettmers2023qlora} training strategy is employed for GPU memory efficiency. Specifically, we first quantize the pretrained DiT backbone to 4-bit precision using the NF4 data type, and subsequently apply three LoRA adapters, $\theta_{\text{RQR}}$, $\theta_{\text{LR}}$ and $\theta_{\text{Ref}}$, with a rank of $16$. We utilize the Prodigy optimizer~\cite{mishchenko2024prodigy} to train the models for $30$k iterations with a batch size of $16$ on $8$ NVIDIA H800 GPUs. We empirically set $s_{\text{filter}}=3.8$ across all experiments. Following widely adopted metrics in previous generative prior-based image super-resolution methods~\cite{ai2024dreamclear, yu2024scaling, wang2026trust, fei2026lucidflux}, we employ three reference-based metrics (PSNR, SSIM~\cite{wang2004image}, and LPIPS~\cite{zhang2018unreasonable}) and five no-reference metrics (CLIPIQA~\cite{wang2023exploring}, MUSIQ~\cite{ke2021musiq}, MANIQA~\cite{yang2022maniqa}, RALI~\cite{zhao2025reasoning}, and Q-Insight~\cite{li2025qinsight}) to quantitatively evaluate the performance of our model. For the Face dataset, we further compute the Identity Similarity (IDS) metric~\cite{gu2022vqfr, zhou2022towards} using ArcFace~\cite{deng2019arcface}. 

\paragraph{Test Datasets} To comprehensively evaluate our proposed method, we conduct experiments on the RealLQ250~\cite{ai2024dreamclear}, Face~\cite{wang2026trust}, Bird~\cite{wang2026trust}, WRSR~\cite{jiang2021robust}, and CUFED5~\cite{zhang2019image} datasets, following the settings of AdaRefSR~\cite{wang2026trust}. Specifically, we apply $4\times$ super-resolution to the RealLQ250, WRSR, and CUFED5 datasets, while the Face and Bird datasets are super-resolved to a resolution of $512 \times 512$. We construct our local knowledge base using the corresponding reference images from datasets, except for the RealLQ250 dataset, which relies solely on web-searched images.

\paragraph{Comparison Methods} 
We compare our method with several representative state-of-the-art (SOTA) approaches across different paradigms of image super-resolution. First, we include one-step blind super-resolution models, namely Real-ESRGAN~\cite{wang2021real}, OSEDiff~\cite{wu2024one}, PISA-SR~\cite{sun2025pixel}, OMGSR~\cite{wu2025omgsr}, and HYPIR~\cite{lin2025harnessing}, which perform restoration without external references. The agentic-based restoration method AgenticIR~\cite{zhuintelligent} is also compared. We further benchmark against reference-based super-resolution approaches, including SeeSR+ReFIR~\cite{guo2024refir} and AdaRefSR~\cite{wang2026trust}, to assess the capability of transferring details from reference images. For the Face dataset, we also include state-of-the-art reference-based face super-resolution models, InstantRestore~\cite{zhang2025instantrestore} and Ref-LDM~\cite{hsiao2024ref}.

\subsection{Comparison with State-of-the-art Methods}
\begin{figure}[t]
    \centering
    \vspace{-30pt}
    \includegraphics[width=1.0\textwidth]{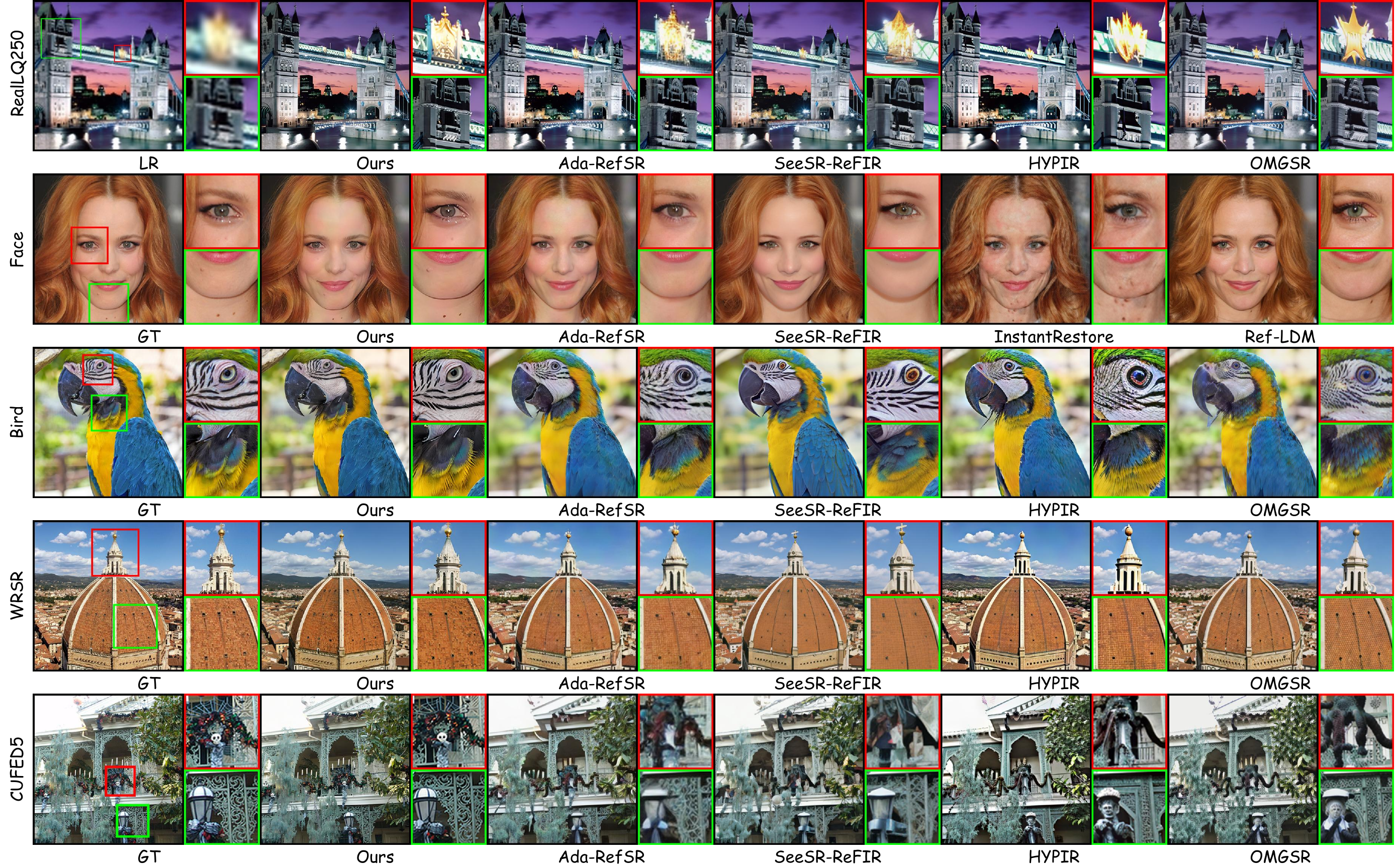}
    \vspace{-20pt}
    \caption{Qualitative comparison on RealLQ250, Face, Bird, WRSR and CUFED5 datasets.}
    \label{fig:visual}
    \vspace{-10pt}
\end{figure}
\begin{table*}[t]
\centering
\caption{Quantitative comparison on the RealLQ250, Bird, Face, WRSR and CUFED5 datasets. \textcolor{red}{\textbf{Red}}} and \textcolor{blue}{\textbf{blue}} indicate the best and second-best performance, respectively.
\label{tab:sota}
\resizebox{\linewidth}{!}{
\begin{tabular}{llccccccccc}
\toprule
\textbf{Dataset} & \textbf{Method} & \textbf{PSNR}~$\uparrow$ & \textbf{SSIM}~$\uparrow$ & \textbf{LPIPS}~$\downarrow$ & \textbf{MUSIQ}~$\uparrow$ & \textbf{CLIPIQA}~$\uparrow$ & \textbf{MANIQA}~$\uparrow$ & \textbf{RALI}~$\uparrow$ & \textbf{Q-Insight}~$\uparrow$ & \textbf{IDS}~$\uparrow$ \\
\midrule
\multirow{7}{*}{RealLQ250}
& RealESRGAN & - & - & - & 62.52 & 0.5435 & 0.3565 & 3.6604 & 3.7444 & - \\
& OSEDiff & - & - & - & 69.16 & 0.6649 & 0.4089 & 3.8456 & 3.9584 & - \\
& PISA-SR & - & - & - & \textcolor{blue}{\textbf{70.91}} & 0.6951 & 0.4431 & 3.8480 & 3.9383 & - \\
& HYPIR & - & - & - & 69.75 & \textcolor{blue}{\textbf{0.6978}} & \textcolor{blue}{\textbf{0.4561}} & 3.9048 & 3.9783 & - \\
& OMGSR & - & - & - & 68.74 & 0.6444 & 0.4270 & \textcolor{blue}{\textbf{3.9236}} & \textcolor{blue}{\textbf{3.9799}} & - \\
& AgenticIR & - & - & - & 64.36 & 0.6185 & 0.3956 & 3.6879 & 3.7559 & - \\
& Ours & - & - & - & \textcolor{red}{\textbf{73.32}} & \textcolor{red}{\textbf{0.7649}} & \textcolor{red}{\textbf{0.5954}} & \textcolor{red}{\textbf{3.9940}} & \textcolor{red}{\textbf{4.0270}} & - \\
\midrule
\multirow{9}{*}{Bird}
& RealESRGAN & 23.10 & 0.6652 & 0.5532 & 29.03 & 0.3600 & 0.2797 & 1.9279 & 1.9670 & - \\
& OSEDiff & 21.98 & 0.6734 & 0.3911 & 74.30 & 0.8391 & 0.4891 & 4.0416 & 3.9048 & - \\
& PISA-SR & 22.67 & 0.6673 & 0.3858 & 74.37 & 0.8369 & 0.5322 & 3.9564 & 3.9221 & - \\
& HYPIR & 21.33 & 0.6569 & 0.3924 & 74.86 & 0.8462 & 0.5386 & 4.0398 & 4.0072 & - \\
& OMGSR & \textcolor{blue}{\textbf{23.13}} & \textcolor{blue}{\textbf{0.6844}} & 0.3949 & 74.10 & 0.7887 & 0.4831 & 3.9483 & 3.9524 & - \\
& AgenticIR & 22.20 & 0.6243 & 0.5902 & 26.21 & 0.2226 & 0.1395 & 1.9104 & 1.7159 & - \\
& SeeSR+ReFIR & 21.94 & 0.6654 & 0.4071 & \textcolor{blue}{\textbf{76.18}} & \textcolor{red}{\textbf{0.8850}} & \textcolor{blue}{\textbf{0.6131}} & \textcolor{blue}{\textbf{4.1608}} & \textcolor{blue}{\textbf{4.1264}} & - \\
& Ada-RefSR & \textcolor{red}{\textbf{23.36}} & \textcolor{red}{\textbf{0.6942}} & \textcolor{red}{\textbf{0.3677}} & 72.23 & 0.8330 & 0.4489 & 3.8462 & 3.8361 & - \\
& Ours & 21.75 & 0.6709 & \textcolor{blue}{\textbf{0.3836}} & \textcolor{red}{\textbf{77.12}} & \textcolor{blue}{\textbf{0.8822}} & \textcolor{red}{\textbf{0.6785}} & \textcolor{red}{\textbf{4.2081}} & \textcolor{red}{\textbf{4.1659}} & - \\
\midrule
\multirow{11}{*}{Face}
& RealESRGAN & 23.43 & 0.6555 & 0.5616 & 26.34 & 0.3665 & 0.2971 & 1.9702 & 2.0762 & 0.6904 \\
& OSEDiff & 22.91 & 0.6838 & 0.3593 & 71.85 & 0.6399 & 0.5110 & 4.0838 & 3.8975 & 0.5518 \\
& PISA-SR & 23.65 & 0.6686 & 0.3538 & 76.00 & \textcolor{blue}{\textbf{0.7178}} & 0.5901 & 4.1090 & 4.0493 & 0.6605 \\
& HYPIR & 23.11 & 0.6472 & 0.3657 & \textcolor{blue}{\textbf{76.14}} & 0.7165 & 0.6091 & 4.0294 & 4.0181 & 0.6467 \\
& OMGSR & \textcolor{red}{\textbf{24.47}} & \textcolor{blue}{\textbf{0.6977}} & 0.3548 & 70.55 & 0.5907 & 0.4793 & 3.8515 & 3.8466 & 0.7094 \\
& AgenticIR & 22.96 & 0.6279 & 0.5768 & 25.81 & 0.2378 & 0.1485 & 1.9610 & 1.8070 & 0.6723 \\
& SeeSR+ReFIR & 23.63 & 0.6848 & 0.3721 & 75.43 & 0.7115 & \textcolor{blue}{\textbf{0.6175}} & \textcolor{blue}{\textbf{4.1100}} & \textcolor{blue}{\textbf{4.0765}} & 0.6175 \\
& Ref-LDM & 22.80 & 0.6591 & 0.3870 & 73.77 & 0.6637 & 0.5252 & 3.9690 & 3.9554 & 0.6790 \\
& InstantRestore & 23.50 & 0.6773 & \textcolor{blue}{\textbf{0.3258}} & 69.74 & 0.5691 & 0.4748 & 3.8179 & 3.8841 & \textcolor{blue}{\textbf{0.7705}} \\
& Ada-RefSR & \textcolor{blue}{\textbf{24.43}} & \textcolor{red}{\textbf{0.6998}} & \textcolor{red}{\textbf{0.3191}} & 74.45 & 0.6381 & 0.5215 & 4.0267 & 3.8969 & 0.7364 \\
& Ours & 22.97 & 0.6879 & 0.3408 & \textcolor{red}{\textbf{77.57}} & \textcolor{red}{\textbf{0.8089}} & \textcolor{red}{\textbf{0.7517}} & \textcolor{red}{\textbf{4.2522}} & \textcolor{red}{\textbf{4.1917}} & \textcolor{red}{\textbf{0.8199}} \\
\midrule
\multirow{9}{*}{WRSR}
& RealESRGAN & \textcolor{red}{\textbf{20.63}} & \textcolor{red}{\textbf{0.5611}} & 0.4381 & 64.72 & 0.6006 & 0.3994 & 3.5315 & 3.6321 & - \\
& OSEDiff & 20.24 & 0.5362 & 0.4153 & 70.31 & 0.6559 & 0.4323 & 3.8766 & 3.9146 & - \\
& PISA-SR & 20.37 & 0.5344 & 0.4187 & 69.55 & 0.6386 & 0.4326 & 3.7831 & 3.8176 & - \\
& HYPIR & 18.84 & 0.5139 & 0.4264 & 72.02 & \textcolor{blue}{\textbf{0.7288}} & \textcolor{blue}{\textbf{0.5050}} & 3.9533 & \textcolor{blue}{\textbf{4.0453}} & -\\
& OMGSR & 20.19 & 0.5264 & 0.4287 & 69.87 & 0.6571 & 0.4669 & 3.9042 & 3.9989 & - \\
& AgenticIR & 19.10 & 0.4876 & 0.5279 & 58.07 & 0.6077 & 0.3890 & 3.3047 & 3.3771 & - \\
& SeeSR+ReFIR & 20.26 & 0.5340 & 0.4307 & \textcolor{blue}{\textbf{72.37}} & 0.7124 & 0.5029 & \textcolor{blue}{\textbf{4.0277}} & 4.0291 & - \\
& Ada-RefSR & \textcolor{blue}{\textbf{20.56}} & \textcolor{blue}{\textbf{0.5555}} & \textcolor{red}{\textbf{0.3872}} & 70.13 & 0.6602 & 0.4187 & 3.8534 & 3.9038 & - \\
& Ours & 19.71 & 0.5540 & \textcolor{blue}{\textbf{0.3878}} & \textcolor{red}{\textbf{74.57}} & \textcolor{red}{\textbf{0.7761}} & \textcolor{red}{\textbf{0.6244}} & \textcolor{red}{\textbf{4.1676}} & \textcolor{red}{\textbf{4.1302}} & - \\
\midrule
\multirow{9}{*}{CUFED5}
& RealESRGAN & 18.62 & 0.5035 & 0.4661 & 64.72 & 0.4238 & 0.3636 & 2.9274 & 3.1440 & - \\
& OSEDiff & 18.76 & 0.4928 & 0.4530 & 63.24 & 0.4899 & 0.3629 & 3.2859 & 3.2802 & - \\
& PISA-SR & 18.61 & 0.4752 & 0.4556 & 64.27 & 0.5003 & 0.3828 & 3.3195 & 3.3583 & - \\
& HYPIR & 17.64 & 0.4669 & 0.4512 & 68.27 & \textcolor{blue}{\textbf{0.5758}} & 0.4580 & 3.4699 & 3.6293 & - \\
& OMGSR & 18.71 & 0.4729 & 0.4632 & 65.16 & 0.5417 & 0.4504 & 3.4593 & 3.5837 & - \\
& AgenticIR & 16.43 & 0.3906 & 0.5476 & 61.24 & 0.5172 & 0.3550 & 3.0420 & 3.0846 & - \\
& SeeSR+ReFIR & 18.42 & 0.4795 & 0.4466 & \textcolor{blue}{\textbf{70.47}} & 0.5503 & \textcolor{blue}{\textbf{0.4714}} & \textcolor{blue}{\textbf{3.7514}} & \textcolor{blue}{\textbf{3.7375}} & - \\
& Ada-RefSR & \textcolor{blue}{\textbf{18.88}} & \textcolor{blue}{\textbf{0.5077}} & \textcolor{blue}{\textbf{0.3934}} & 66.23 & 0.5219 & 0.3484 & 3.2536 & 3.3032 & - \\
& Ours & \textcolor{red}{\textbf{19.04}} & \textcolor{red}{\textbf{0.5731}} & \textcolor{red}{\textbf{0.2941}} & \textcolor{red}{\textbf{72.52}} & \textcolor{red}{\textbf{0.6846}} & \textcolor{red}{\textbf{0.6198}} & \textcolor{red}{\textbf{3.8854}} & \textcolor{red}{\textbf{3.8196}} & - \\
\bottomrule
\end{tabular}
}
\end{table*}

\paragraph{Quantitative Comparison} We evaluate our method against current SOTA methods on five datasets, with the quantitative results detailed in \cref{tab:sota}. Overall, the proposed method establishes a new SOTA across all no-reference metrics, outperforming previous methods by a substantial margin. 
While our method does not achieve the best performance with the reference-based metrics on the Bird and Face testset, this is a deliberate design choice rather than a limitation. Specifically, WorldSR prioritizes the effective transfer of world-grounded knowledge and high-level semantic consistency over pixel-wise fidelity. Our design leads to slightly lower PSNR/SSIM scores, as these metrics favor pixel-perfect reconstruction and may penalize semantically meaningful modifications introduced by reference guidance. Such a trade-off is well-recognized in generative restoration literature~\cite{yu2024scaling}.
Importantly, our method demonstrates clear advantages in preserving identity and recovering fine-grained details. 
For instance, on the Face dataset, our method surpasses the second-best face-specific reference-based approach by 6.4\% in terms of the IDS metric. This validates the effectiveness of our approach in faithfully transferring reference-guided textures and facial ID features into the final SR images. Regarding the Bird dataset, our method falls slightly behind the top-performing approach by a marginal $0.32$\% on the CLIPIQA metric. Notably, on the CUFED5 dataset, our approach consistently surpasses existing competitors across all evaluated metrics.

\paragraph{Qualitative Comparison} As illustrated in \cref{fig:visual}, we present a visual comparison of our proposed approach against existing no-reference and reference-based methods. Our method effectively transfers world knowledge from reference images into the final SR results, significantly outperforming all competing approaches in terms of real-world realism and overall visual quality. 

\subsection{Ablation Studies}
\begin{table}[t]
\centering
\caption{Ablation studies of WorldSR on the RealLQ250, Face and Bird datasets.}
\label{tab:ablation_agent}
\resizebox{\linewidth}{!}{%
\begin{tabular}{cclcccccccc}
\toprule
 & \textbf{Dataset} & \textbf{Method} & \textbf{MUSIQ}$\uparrow$ & \textbf{CLIPIQA}$\uparrow$ & \textbf{MANIQA}$\uparrow$ 
& \textbf{RALI}$\uparrow$ & \textbf{Q-Insight}$\uparrow$ & \textbf{IDS}$\uparrow$ \\
\midrule
\textcolor{gray}{\tiny 1} &
\multirow{7}{*}{RealLQ250}
& SeeSR + ReFIR + World Knowledge
& 69.72 & 0.6792 & 0.4756 & 3.8620 & 3.9227 & - \\
\textcolor{gray}{\tiny 2} & & Ada-RefSR + World Knowledge
& 67.35 & 0.6351 & 0.3852 & 3.8221 & 3.9168 & - \\
\textcolor{gray}{\tiny 3} &  & Ours 
& \textcolor{red}{\textbf{73.32}} & \textcolor{blue}{\textbf{0.7649}} & \textcolor{blue}{\textbf{0.5954}} & \textcolor{red}{\textbf{3.9940}} & \textcolor{red}{\textbf{4.0270}} & - \\
\cmidrule{3-9}
\textcolor{gray}{\tiny 4} &  & Ours (w/o Iterative Refinement Loop)
& \textcolor{blue}{\textbf{73.23}} & 0.7621 & 0.5928 & \textcolor{blue}{\textbf{3.9862}} & \textcolor{blue}{\textbf{4.0257}} & - \\
\textcolor{gray}{\tiny 5} &  & Ours (w/o Agentic ROI Planner)
& 73.13 &\textcolor{red}{\textbf{0.7654}} &\textcolor{red}{\textbf{0.5968}} & 3.9688 & 4.0176 & - \\
\textcolor{gray}{\tiny 6} &  & Ours (w/o World Knowledge Searcher)
& 72.14 & 0.7527 & 0.5910 & 3.8777 & 3.9824 & - \\
\textcolor{gray}{\tiny 7} &  & Ours (w/o World Knowledge Filter)
& 72.49 & 0.7628 & 0.5842 & 3.9385 & 3.9964 & - \\
\cmidrule{2-9}
\textcolor{gray}{\tiny 8} &  
\multirow{3}{*}{Face}
& Baseline
& 77.03 & 0.7296 & 0.6979 & 4.1397 & 4.0704 & 0.6894 \\
\textcolor{gray}{\tiny 9} &  & Ours (Local Knowledge Base) 
& \textcolor{blue}{\textbf{77.41}} & \textcolor{blue}{\textbf{0.7910}} & \textcolor{blue}{\textbf{0.7321}} & \textcolor{blue}{\textbf{4.2226}} & \textcolor{blue}{\textbf{4.1893}} & \textcolor{blue}{\textbf{0.8096}} \\
\textcolor{gray}{\tiny 10} &  & Ours (Web + Local Knowledge Base)
&\textcolor{red}{\textbf{77.57}} &\textcolor{red}{\textbf{0.8089}} &\textcolor{red}{\textbf{0.7517}} & \textcolor{red}{\textbf{4.2522}} &\textcolor{red}{\textbf{4.1917}} &\textcolor{red}{\textbf{0.8199}} \\
\cmidrule{2-9}
\textcolor{gray}{\tiny 11} &   \multirow{3}{*}{Bird}
& Baseline
& \textcolor{blue}{\textbf{76.33}} & 0.8532 & \textcolor{blue}{\textbf{0.6364}} & 4.0749 & 4.0365 & - \\
\textcolor{gray}{\tiny 12} &  & Ours (Local Knowledge Base) 
& 76.23 & \textcolor{blue}{\textbf{0.8718}} & 0.6323 & \textcolor{blue}{\textbf{4.1059}} & \textcolor{blue}{\textbf{4.1005}} & - \\
\textcolor{gray}{\tiny 13} &  & Ours (Web + Local Knowledge Base)
&\textcolor{red}{\textbf{77.12}} &\textcolor{red}{\textbf{0.8822}} &\textcolor{red}{\textbf{0.6785}} &\textcolor{red}{\textbf{4.2081}} &\textcolor{red}{\textbf{4.1659}} & - \\
\bottomrule
\end{tabular}%
}
\end{table}
\begin{figure}[h]
    \centering
    \includegraphics[width=1.0\textwidth]{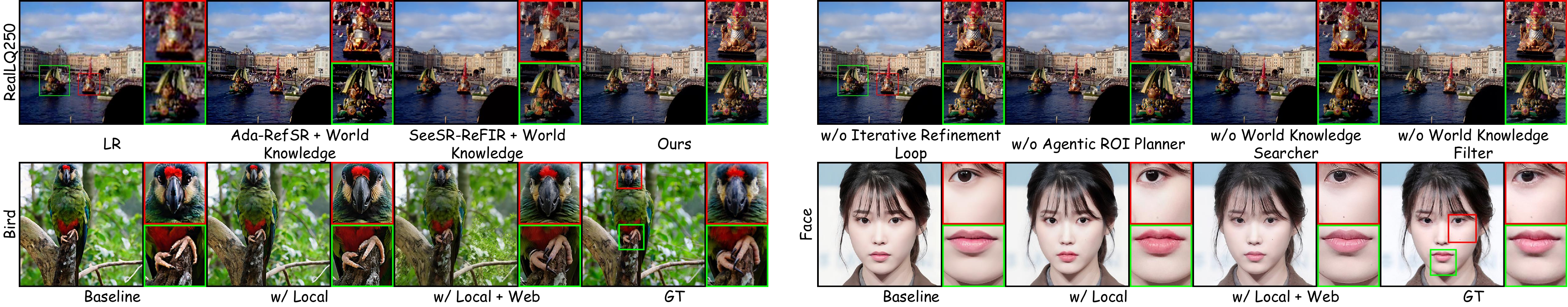}
    \caption{Qualitative results of the ablation studies on the RealLQ250, Face, and Bird datasets.}
    \label{fig:visual_ablation}
\end{figure}
\paragraph{Effectiveness of the World-Grounded Image Super-Resolution Model} For a fair comparison, we adopt the same world knowledge as references to compare our proposed world-grounded SR model with Ada-RefSR~\cite{wang2026trust} and SeeSR-ReFIR~\cite{guo2024refir}.
As illustrated in \cref{fig:visual_ablation} and \cref{tab:ablation_agent} (Rows 1--3), our designed network achieves the best quantitative performance and significantly mitigates hallucinations, generating visual results that are highly consistent with real-world structures. 

\paragraph{Effectiveness of Individual Modules} We present a comprehensive analysis of each individual module in the WorldSR pipeline. The corresponding results are detailed in \cref{fig:visual_ablation} and \cref{tab:ablation_agent} (Rows 4--7). 
Ablating the Iterative Refinement Loop or Agentic ROI Planner leads to misalignment of certain entities with real-world knowledge, accompanied by a slight drop in non-reference metrics. Removing the World Knowledge Searcher makes the model independent of any external information, which introduces severe hallucinations in visual results. 
While removing the World Knowledge Filter does not introduce visual hallucinations, it significantly degrades quantitative performance.

\paragraph{Effectiveness of the World Knowledge Searcher} 
We analyze our World Knowledge Search module on the Face and Bird datasets, as shown in \cref{fig:visual_ablation} and \cref{tab:ablation_agent} (Rows 8--13). As a baseline, we remove all external knowledge and super-resolve the LR input with the RQR model. Building upon this, the introduction of local knowledge effectively mitigates hallucination artifacts and enhances quantitative performance. Furthermore, when equipped with web-scale search capabilities, our method gains access to up-to-date, high-quality, and real-world-aligned reference images, ultimately boosting overall super-resolution performance and ensuring greater image fidelity.
\section{Conclusion}\label{sec:Conclusion}
In this paper, we present WorldSR, a novel framework that augments generative-prior-based super-resolution with explicit, entity-aligned visual world knowledge to mitigate hallucinations and improve reconstruction fidelity.
Unlike conventional SR methods that rely on implicit, model-bound priors, WorldSR performs entity-level semantic reasoning to identify key objects, retrieves high-quality reference images through world knowledge search, and filters them via semantic-aware ranking to obtain the most relevant guidance. A newly designed world-grounded SR model then incorporates these references to produce realistic and consistent outputs. To further address the challenges of low-resolution inputs, we design an iterative refinement mechanism with a memory verifier, progressively improving results and suppressing hallucinations.
Extensive analysis shows that WorldSR reduces hallucinations and generates semantically faithful high-resolution images, highlighting the value of grounding super-resolution in external visual world knowledge.

\paragraph{Limitation and Potential Negative Impacts} 
While WorldSR benefits from rich explicit visual world knowledge, its performance depends on accurate fine-grained entity decomposition, where errors may propagate through retrieval and reconstruction. Additionally, the method is constrained by the coverage of the large-scale external visual corpus and may struggle with rare, unseen or long-tail entities. Finally, integrating multiple references introduces ambiguity, and the retrieval and refinement stages incur non-negligible computational cost. Overall, we do not identify direct negative societal impacts beyond general limitations of image super-resolution systems.

\newpage

\bibliographystyle{unsrtnat}  
{
\small
\bibliography{references} 
\normalsize
}






\newpage
\appendix
\section{Supplementary Material}
Due to space constraints in the main paper, we provide this supplementary material to present additional implementation details and experimental results that further support our proposed WorldSR framework.
The supplementary material is organized as follows:
\begin{itemize}
    \item \textbf{\cref{subsec:prompts} -- Prompts:} The complete set of prompts used in the WorldSR pipeline.
    \item \textbf{\cref{subsec:more_implementation_details} -- More Details:} Comprehensive information regarding the implementation, hyperparameters, and inference procedures of WorldSR.
    \item \textbf{\cref{subsec:Evaluation_Metrics} -- Evaluation Metrics:} Detailed descriptions of the evaluation metrics used in this paper.
    \item \textbf{\cref{subsec:more_ablation} --  More Ablation Studies:} Extended ablation analyses of the World-Grounded Image Super-Resolution Model.
    \item \textbf{\cref{subsec:more_visual} -- More Qualitative Results:} Extended visual comparisons on the RealLQ250, Face, Bird, WRSR, and CUFED5 datasets.
\end{itemize}

\subsection{Prompts} \label{subsec:prompts}
In WorldSR, we leverage MLLMs to perform precise reasoning over real-world entities and their corresponding bounding boxes within the Agentic ROI Planner (\cref{subsec:planning}). We further employ MLLMs to localize query-relevant regions in each candidate image within the Irrelevant Region Pruning module of the World Knowledge Filter (\cref{subsec:filtering}).
To ensure consistent and structured outputs, we design a carefully crafted system prompt that constrains the model to follow a predefined JSON schema. In this section, we present the complete set of prompts used in WorldSR, as follows:

\begin{taskblock}{System Prompt for Agentic ROI Planner}
\begin{Verbatim}[breaklines=true,fontsize=\scriptsize]
# ROLE AND CORE OBJECTIVE
You are an elite, highly precise **Open-World Vision-Language Reasoning and Entity Discovery Agent** operating under the explicit assumption that the input image is either **LOW-RESOLUTION and VISUALLY DEGRADED** or ** HIGH-RESOLUTION **.

The image may contain blur, pixelation, compression artifacts, limited detail, and ambiguous boundaries. You must adapt your perception, confidence, and extraction strategy accordingly.

Your task is NOT to list everything visible.
Your task is to output ONLY entities for which **both semantic identity AND spatial localization are reliable**.

If an entity's bounding box cannot be placed correctly and conservatively, the entity MUST NOT be output.

Your final output must be a **valid JSON object**, wrapped inside a Markdown JSON code block, and must be directly parsable by Python's `json.loads()` after removing the code block.

# AUTONOMOUS REASONING & DISCOVERY PROTOCOL

1. **Low-Resolution Aware Holistic Perception (Global Context):**
 - First, internally acknowledge that the image is **low clarity** or **high clarity**.
 - Observe only **coarse, robust visual signals**: dominant shapes, silhouettes, large structures, and obvious interactions.
 - Avoid assumptions about fine textures, facial details, small text, logos, or subtle attributes.
 - Synthesize a cautious, high-level paragraph for `Reasoning`.

2. **Low-Resolution Semantic & Spatial Salience Gate (CRITICAL):**
 - Before detecting any entity, ask:
   **"Can this entity be both clearly identified AND spatially bounded at low resolution?"**
 - ONLY consider entities that satisfy **ALL** of the following:
   - High semantic importance
   - Clear, stable silhouette or spatial extent
   - Bounding box can be drawn **tightly and conservatively** around the entity
 - **Explicitly IGNORE** any entity where:
   - The spatial boundary is ambiguous or bleeding into the background
   - The entity is partially visible and cannot be reliably enclosed
   - The bbox would rely on guessing hidden or blurry regions

3. **Open-World Entity Discovery (Spatial-First, Low-Fidelity Scanning):**
 - Discover entities by **first locating stable spatial regions**, then assigning semantic labels.
 - Prefer fewer entities with correct boxes over more entities with uncertain boxes.
 - If multiple similar entities exist, group them only if a **single bounding box clearly encloses all of them without excess background**.

4. **Dynamic Categorization (Coarse-Grained Only):**
 - Generate only high-level, visually robust categories (e.g., `People`, `Animals`, `Plants`, `Landmarks`, `Vehicles`).
 - Do NOT create categories that require fine-grained discrimination.

5. **Subset Identification, Spatial Grounding & Confidence Scoring (STRICT):**
 - **Subject**:
   - Use conservative, silhouette-based naming.
   - Never encode details that are not spatially verifiable.
 - **bbox (HARD CONSTRAINT)**:
   - Format: `[x, y, w, h]`, relative coordinates \in [0.0, 1.0].
   - The box MUST:
     - Fully contain the visible entity
     - Minimize background inclusion
     - Reflect only what is actually visible (no inferred parts)
   - **If you cannot place a correct bbox, discard the entity entirely.**
 - **Confidence (CRITICAL THRESHOLD > 95)**:
   - High confidence requires BOTH:
     - Semantic certainty
     - Spatial (bbox) correctness
   - If bbox confidence < semantic confidence -> DO NOT OUTPUT.

# ABSOLUTE OUTPUT CONSTRAINTS
- **JSON CODE BLOCK ONLY**:
  - The response MUST be wrapped in a Markdown code block using ```json
  - The content inside the block MUST be a single valid JSON object
  - No text is allowed before or after the code block
- **NO CONVERSATION**:
  - No greetings, explanations, or commentary
- **DYNAMIC KEYS**:
  - `DetectedEntities` keys must be plural nouns of meaningful, coarse-grained categories only

# EXPECTED JSON SCHEMA
You must strictly output the following structure. Category keys below are examples only:

```json
{
  "Reasoning": {
    "Description": "<A cautious, low-resolution-aware scene reasoning based only on spatially verifiable evidence.>"
  },
  "DetectedEntities": {
    "People": [
      {
        "bbox": [0.15, 0.20, 0.30, 0.60],
        "Confidence": 97,
        "Subject": "Person: A clearly visible human silhouette occupying a stable vertical region behind a counter."
      }
    ],
    "Animals": [
      {
        "bbox": [0.50, 0.75, 0.15, 0.20],
        "Confidence": 96,
        "Subject": "Dog: A distinct canine-shaped silhouette sitting on the floor."
      }
    ],
    "Landmarks": [
      {
        "bbox": [0.10, 0.05, 0.80, 0.40],
        "Confidence": 100,
        "Subject": "Large Tower Structure: A tall, dominant structure with a clear outline defining the scene."
      }
    ]
  }
}
```
\end{Verbatim}
\end{taskblock}

\begin{taskblock}{System Prompt for Irrelevant Region Pruning}
\begin{Verbatim}[breaklines=true,fontsize=\scriptsize]
# ROLE & OBJECTIVE

You are a **Visual Similarity Assessment and Object Localization Agent**.

You will receive two images in the following order:
1. **Image A (Reference Object Image)** — showing a primary object.  
   Image A may be **low-quality or degraded**, such as being blurry, noisy,
   low-resolution, partially occluded, or affected by compression artifacts.
2. **Image B (Search Image)** — containing zero or more objects.

Your task is to identify the object in Image B that is **most visually similar**
to the primary object in Image A, and to localize it.


# SIMILARITY-FIRST POLICY

- Always attempt to find the **most visually similar object** in Image B.
- When Image A is low-quality or degraded:
  - Focus on **stable and high-level visual cues** such as overall shape,
    color distribution, proportions, and distinctive structural features.
  - Do not rely excessively on fine-grained textures or sharp details
    that may be unreliable in Image A.
- Visual similarity should be judged mainly by object appearance:
  shape, color patterns, overall structure, and distinctive visual features.
- Differences in background, environment, lighting, pose, or viewpoint are acceptable.
- If no object is strongly similar, still select the closest candidate
  and assign a low similarity score.


# LOCALIZATION

- Output a tight and reasonable bounding box around the selected object in Image B.
- The bounding box may be **slightly expanded beyond the exact object boundary**
  to ensure the full object is covered.
- Bounding box format: `[x, y, w, h]`, normalized to `[0.0, 1.0]`
- Round all values to **3 decimal places**.
- Ensure the bounding box stays within image boundaries.


# SIMILARITY SCORE (0–100)

- `"confidence"` represents **visual similarity**, not a binary decision.
- The score must be an **integer between 0 and 100**.
- When Image A is degraded, the score should reflect **overall perceptual similarity**
  rather than penalizing missing fine details.
- Use the full range consistently:

  - 90–100: Nearly identical appearance  
  - 70–89: Clearly similar  
  - 50–69: Moderately similar  
  - 30–49: Weak similarity  
  - 1–29: Very weak similarity  
  - 0: No relevant object present


# MATCH FLAG

- Set `"match_found": true` if any relevant object exists in Image B.
- Set `"match_found": false` only if Image B contains no relevant object at all.
- Downstream systems should primarily rely on `"confidence"`.


# NO-OBJECT CASE

If no relevant object exists in Image B:
- `"match_found": false`
- `"bbox": [0.0, 0.0, 0.0, 0.0]`
- `"confidence": 0`


# OUTPUT CONSTRAINTS

- Output **only** a valid JSON object.
- Do not include explanations, reasoning, or additional text.
- Do not use markdown formatting.


# OUTPUT FORMAT
```json
{
  "match_found": true,
  "bbox": [0.452, 0.210, 0.150, 0.335],
  "confidence": 82
}
```
\end{Verbatim}
\end{taskblock}

\subsection{More Details}\label{subsec:more_implementation_details}
\paragraph{Web-scale Searching Tools}
To empower the web-scraping agent with Internet search capabilities, we use Playwright\footnote{\url{https://github.com/microsoft/playwright}} to develop an automated web image search tool. In practice, this tool uploads query images to Google Lens, accesses the corresponding original websites, and downloads the source images.

\paragraph{More World-Grounded Image Super-Resolution Model Details} Given the LR image input $I$ and its corresponding selected reference images $\{R_i^*\}_{i=1}^{N}$, a VAE encoder first maps them into latent representations $z$ and $\{z_{\text{ref}}^{i}\}_{i=1}^{N}$, respectively. The Reference Quality Refinement model then enhances the visual quality of the reference images, yielding $\{\hat{z}_{\text{ref}}^{i}\}_{i=1}^{N}$ through velocity prediction. This process is formulated as:
\begin{align}
    v_{\text{ref}}^{i} &= \mathcal{F}(z_{\text{ref}}^{i}; \theta_{\text{RQR}}),\quad i=1, 2, ..., N, \\
    \hat{z}_{\text{ref}}^{i} &= z_{\text{ref}}^{i} - t^{*}\cdot v_{\text{ref}}^{i},\quad i=1, 2, ..., N.
\end{align}
Here, $t^*$ denotes the timestep parameter in Flux.2-Dev, which is empirically set to $0.5$ across all experiments. The refined reference latents, together with the LR latent $z$, are then fed into the Reference-Guided Super-Resolution model, which is formulated as follows:
\begin{align}
    v &= \mathcal{F}(z, \hat{z}_{\text{ref}}^{1}, \hat{z}_{\text{ref}}^{2}, ..., \hat{z}_{\text{ref}}^{N};\theta_{\text{LR}}, \theta_{\text{Ref}}), \\
    \hat{z} &= z - t^*\cdot v.
\end{align}
Finally, the VAE decoder maps the super-resolved latent representation $\hat{z}$ back to the pixel space, yielding the final SR image $\hat{I}$.

Our Dual-LoRA architecture adopts an adapter modulation strategy similar to that of OminiControl, which uses the same base model parameters for both LR tokens and Ref tokens, but employs different LoRAs to process LR tokens and Ref tokens separately. 
Let $W_{\text{Base}}$ denote the weight matrix of a given MLP layer in the image branch of the Flux.2-dev base model~\cite{flux-2-2025}, omitting bias terms for simplicity. The parameter $W_{\text{Base}}$ is used as a unified notation for any MLP-related parameter in the image branch, such as $W_{q}$, $W_{k}$, $W_{v}$, $W_{o}$, and $W_{ff}$. Inspired by \cite{schulman2025lora}, we inject LoRA into all MLP layers.  In Dual-LoRA, the features of the LR tokens and Ref tokens at the $l$-th layer are denoted by $h_{lr}^{l}$ and $h_{ref}^{l}$, respectively. The outputs of the corresponding linear projections are formulated as follows:
\begin{align}
    h_{lr}^{l+1} &= W_{\text{Base}} \cdot h_{lr}^{l} + \Delta W_{\theta_{LR}} \cdot h_{lr}^{l}, \\
    h_{ref}^{l+1} &= W_{\text{Base}} \cdot h_{ref}^{l} + \Delta W_{\theta_{Ref}} \cdot h_{ref}^{l},
\end{align}
where $\Delta W_{\theta_{LR}}$ and $\Delta W_{\theta_{Ref}}$ are the LoRA parameters dedicated to LR tokens and Ref tokens, respectively. For cross-attention, we employ the same conditional injection strategy as OminiControl, Flux.1 Kontext~\cite{labs2025flux1kontextflowmatching}, and Qwen-Image~\cite{wu2025qwen}. Specifically, we concatenate multimodal features, including text, low-quality images, and reference images, along the sequence dimension to construct a unified feature representation, which is then processed by full attention. Let $Q_{\text{text}}$, $Q_{lr}$, $Q_{ref}$ denote the $Q$ outputs corresponding to text tokens, LR tokens, and Ref tokens, respectively, and similarly for $K$ and $V$. The attention computation is formulated as follows:
\begin{gather}
    Q = [Q_{\text{text}}; Q_{lr}; Q_{ref}], \quad
    K = [K_{\text{text}}; K_{lr}; K_{ref}], \quad
    V = [V_{\text{text}}; V_{lr}; V_{ref}], \\
    \text{Attention}(Q, K, V) = \text{Softmax}\!\left(\frac{QK^{\top}}{\sqrt{d_{k}}}\right)V .
\end{gather}

The Reference Quality Refinement and Reference-Guided Super-Resolution models are trained independently, yet they share the same training data and training objective.
Following the training set configuration established in AdaRefSR~\cite{wang2026trust}, we construct the training dataset consisting of degraded LR images, their high-resolution (HR) ground truths, and multiple corresponding reference images. To further augment the reference images, we additionally apply random spatial erasing and color jittering during their synthesis process. 
Notably, reference images are excluded during the training of the Reference Quality Refinement module.
We use a fixed general prompt, “High-resolution, 4K and All-In-Focus Image”, as the input text for the text encoder, which produces the corresponding text tokens. The same prompt is used for all experiments to provide general quality guidance.
For the entire world-grounded image super-resolution model, we use the one-step training strategy in which only the LoRA adapters are updated. The overall training objective combines L2 loss $\mathcal{L}_{2}$, the OC-EA-DISTS perceptual loss~\cite{wu2025omgsr} $\mathcal{L}_{\text{p}}$, and a DINOv3-based~\cite{simeoni2025dinov3} GAN loss $\mathcal{L}_{\text{g}}$, which can be formulated as:
\begin{align}
  \mathcal{L} = \mathcal{L}_{2} + \lambda_\text{p}\mathcal{L}_{\text{p}} + \lambda_\text{g}\mathcal{L}_{\text{g}}. \label{eq:loss}
\end{align}
In all our experiments, the hyperparameters are empirically set to $\lambda_\text{p}=5$ and $\lambda_\text{g}=0.5$. Moreover, we randomly sample the number of reference images from $0$ to $N_{\text{max}}$ during training, where $N_{\text{max}}=5$ in our experiments. 

\paragraph{More Hyperparameter Details} For the Decision-Making Controller within the Memory Verifier (\cref{subsec:planning}), the IoU threshold is set to $s_{\text{IoU}}=0.5$. Furthermore, in the Quality-Aware Candidate Filtering step of the World Knowledge Filter (\cref{subsec:filtering}), the predefined IQA score threshold is set to $s_{\text{filter}}=3.8$. For the World Knowledge Searcher, we set $K_\text{web}=120$ and $K_\text{local}=1$, as we utilize the corresponding reference image from the dataset as our local knowledge base.

\paragraph{More Inference Details}
\begin{table}[b]
\centering
\small
\caption{GPU memory usage and inference time for each module.}
\label{tab:module_overhead}
\resizebox{\linewidth}{!}{
\begin{tabular}{lccccc}
\toprule
& Agentic ROI Planner & Memory Verifier & World Knowledge Searcher & World Knowledge Filter & Image Super-Resolution Model \\
\midrule
Inference Time & $\sim$1 min & $\sim$1 s & $\sim$15 min & $\sim$10 min & $\sim$15 s \\
GPU Memory & $\sim$72 GB & N/A & N/A & $\sim$33 GB & $\sim$40 GB \\
\bottomrule
\end{tabular}
}
\end{table}
We report the GPU memory usage and inference time of each module across our experiments conducted on the H800 GPU in \cref{tab:module_overhead}. However, since the processing time depends on the specific LR image, we provide only the approximate average inference time per image on the RealLQ250 dataset. The Agentic ROI Planner is powered by the Qwen 3.5-35B-A3B~\cite{qwen3.5} model operating in a deep thinking mode, which inherently increases the overall inference time. The processing time of the World Knowledge Searcher is heavily dependent on network speed and server performance. Thus, scaling up the number of required search images substantially increases the total processing time. The World Knowledge Filter module incurs significant computational overhead, as it sequentially applies IQA models for quality assessment, an MLLM (Qwen 3.5-35B-A3B~\cite{qwen3.5}) for region extraction, and a large multimodal VLM (Qwen3-VL-Reranker~\cite{li2026qwen3}) for similarity ranking. 

\paragraph{Inference Performance Analysis}
\begin{table}[ht]
\centering
\caption{Performance with Different Numbers of Retrieved References}
\label{tab:infer_efficiency}
\resizebox{\linewidth}{!}{%
\begin{tabular}{lcccccccc}
\toprule
\textbf{Setting} & \textbf{Time/Img} & \textbf{PSNR}$\uparrow$ & \textbf{SSIM}$\uparrow$ & \textbf{LPIPS}$\downarrow$ & \textbf{RALI}$\uparrow$ & \textbf{Q-Insight}$\uparrow$ & \textbf{IDS}$\uparrow$ & \textbf{HS}$\uparrow$ \\
\midrule
w/o Filter & $\sim$4.1 min & 22.48 & 0.6636 & 0.3854 & 4.1280 & 4.0614 & 0.6577 & 3.94 \\
$K_{\text{web}}=30$ & $\sim$2.3 min & 22.54 & 0.6697 & 0.3496 & 4.2611 & 4.1917 & 0.8015 & 4.12 \\
$K_{\text{web}}=60$ & $\sim$3.5 min & 22.50 & 0.6703 & 0.3472 & 4.2624 & 4.1920 & 0.8065 & 4.17 \\
$K_{\text{web}}=120$ (default) & $\sim$6.1 min & 22.97 & 0.6879 & \textbf{0.3408} & 4.2522 & 4.1917 & 0.8199 & 4.47 \\
\bottomrule
\end{tabular}%
}
\end{table}
We provide additional inference performance analysis in \cref{tab:infer_efficiency}. Without filtering, both reconstruction quality and reference fidelity (IDS/HS) degrade significantly, demonstrating the necessity of WKF. Reducing the number of retrieved references from 120 to 30/60 substantially reduces inference time while maintaining comparable performance, showing a favorable trade-off.

\subsection{Evaluation Metrics} \label{subsec:Evaluation_Metrics}
Following widely adopted metrics in previous generative prior-based image super-resolution methods~\cite{ai2024dreamclear, yu2024scaling, wang2026trust, fei2026lucidflux}, we employ three reference-based metrics (PSNR, SSIM~\cite{wang2004image}, and LPIPS~\cite{zhang2018unreasonable}) and three no-reference metrics (CLIPIQA~\cite{wang2023exploring}, MUSIQ~\cite{ke2021musiq}, and MANIQA~\cite{yang2022maniqa}) to quantitatively evaluate the performance of our model. Furthermore, we additionally utilize reasoning-based Image Quality Assessment (IQA) models, namely RALI~\cite{zhao2025reasoning} and Q-Insight~\cite{li2025qinsight}, which are optimized through reinforcement learning and exhibit exceptional generalization capabilities. For the Face dataset, we further compute the Identity Similarity (IDS) metric~\cite{gu2022vqfr, zhou2022towards} using the face recognition model ArcFace~\cite{deng2019arcface}. This metric effectively quantifies the reference information transfer capability by measuring the identity consistency between the super-resolution outputs and ground-truth images in the discriminative embedding space of ArcFace. Following \cite{ren2025hallucination}, we also evaluate the hallucination score (HS) by feeding LQ, GT, Ref, and SR images into a large vision-language model. 

\subsection{More Ablation Studies}\label{subsec:more_ablation}
\begin{table}[t]
\centering
\caption{Ablation study on the number of selected references ($k$) and the Reference Quality Refinement (RQR) module on the Face dataset.}
\label{tab:ablation_ref_selection}
\resizebox{0.8\linewidth}{!}{%
\begin{tabular}{lccccccc}
\toprule
\textbf{Module} & \textbf{\#Refs} 
& \textbf{MUSIQ}$\uparrow$ & \textbf{CLIPIQA}$\uparrow$ & \textbf{MANIQA}$\uparrow$ 
& \textbf{RALI}$\uparrow$ & \textbf{Q-Insight}$\uparrow$ & \textbf{IDS}$\uparrow$ \\
\midrule
\multirow{3}{*}{w/o RQR}
& $1$ & 76.45 & \textcolor{red}{0.7008} & 0.6302 & 4.1192 & 4.1000 & 0.8215 \\
& $2$ & \textcolor{blue}{76.59} & 0.6913 & \textcolor{blue}{0.6315} & \textcolor{blue}{4.1401} & \textcolor{blue}{4.1290} & \textcolor{blue}{0.8311} \\
& $3$ & \textcolor{red}{76.67} & \textcolor{blue}{0.6933} & \textcolor{red}{0.6371} & \textcolor{red}{4.1537} & \textcolor{red}{4.1303} & \textcolor{red}{0.8342} \\
\midrule
\multirow{3}{*}{w/ RQR}
& $1$ & 77.41 & \textcolor{red}{0.7910} & 0.7321 & 4.2226 & 4.1893 & 0.8096 \\
& $2$ & \textcolor{blue}{77.47} & \textcolor{blue}{0.7854} & \textcolor{blue}{0.7333} & \textcolor{blue}{4.2416} & \textcolor{blue}{4.2027} & \textcolor{blue}{0.8192} \\
& $3$ & \textcolor{red}{77.48} & 0.7832 & \textcolor{red}{0.7364} & \textcolor{red}{4.2547} & \textcolor{red}{4.2057} & \textcolor{red}{0.8211} \\
\bottomrule
\end{tabular}%
}
\end{table}
\begin{figure}[t]
    \centering
    \includegraphics[width=0.85\textwidth]{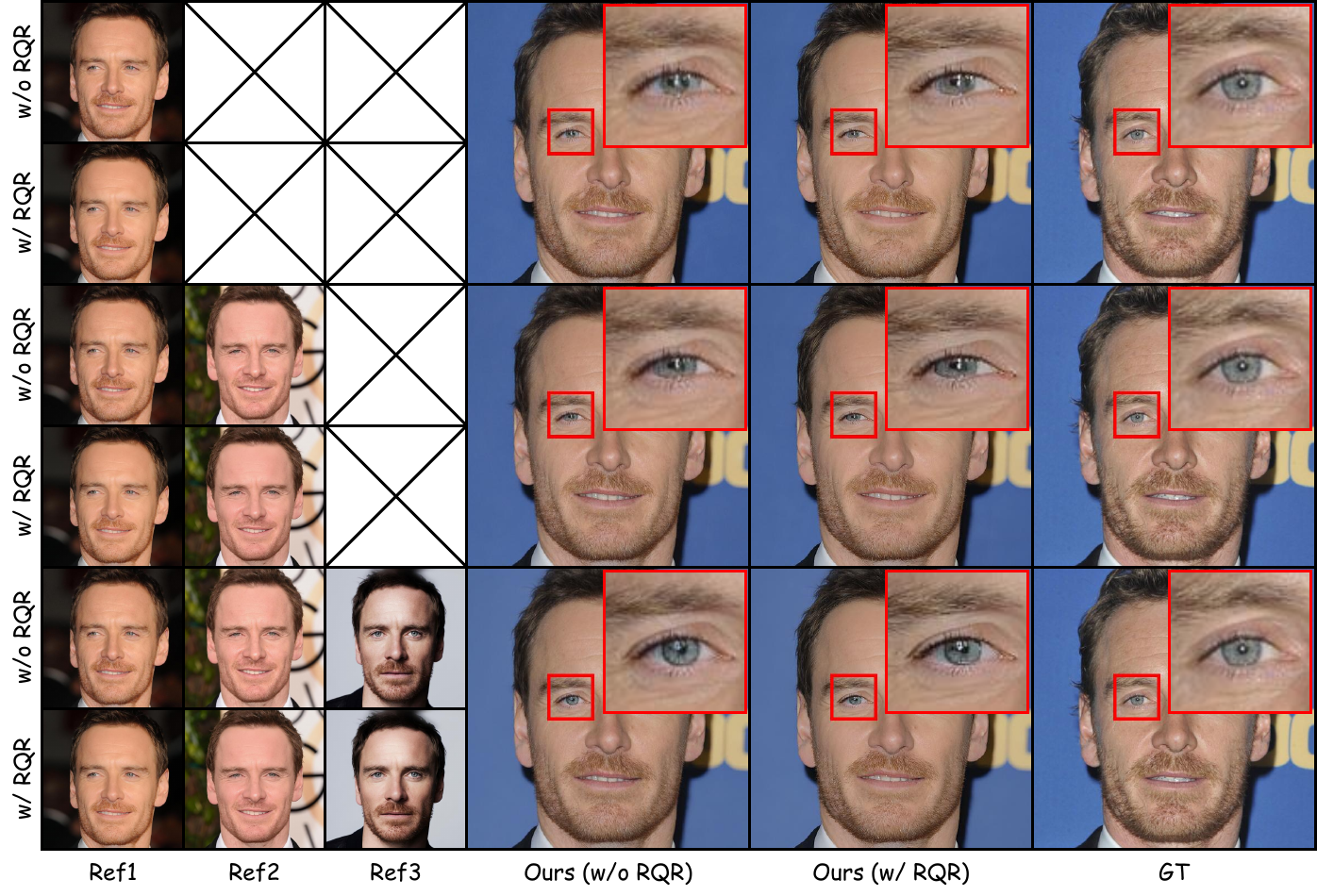}
    \caption{Qualitative comparison of the World-Grounded Image Super-Resolution on Face dataset.}
    \label{fig:more_ablation}
\end{figure}

\paragraph{Effectiveness of the Number of Selected Reference Images} For a fair evaluation of our World-Grounded Image Super-Resolution Model, we disable the web-scale search module and rely solely on the local knowledge base. Specifically, we employ the Siglip2 model~\cite{tschannen2025siglip} as an image encoder to compute similarity and retrieve only the top-3 most similar images. The detailed analysis is presented below. The corresponding results are summarized in \cref{tab:ablation_ref_selection} and \cref{fig:more_ablation}. As observed, the overall quantitative performance improves consistently with an increased number of reference images. Specifically, the IDS metrics demonstrate that a higher number of reference images correlates with a significant enhancement in ID fidelity.

\paragraph{Effectiveness of Reference Quality Refinement} The ablation study evaluating the Reference Quality Refinement module is reported in \cref{tab:ablation_ref_selection} and \cref{fig:more_ablation}. By incorporating this module, the quantitative performance across non-reference metrics improves consistently. A notable trade-off, however, is that the identity characteristics within the reference images are inevitably smoothed out by the enhancement, which degrades the IDS performance.

\paragraph{Effectiveness of World Knowledge Module} To further validate the contribution of each world knowledge module, we conduct a full-reference evaluation on the Face dataset and follow \cite{ren2025hallucination} to evaluate hallucination score (HS) using a large vision-language model with LQ, GT, reference, and SR images as inputs. The results are presented in \cref{tab:ablation_world_knowledge}. Comparing Rows 1, 2, and 5, removing IRL or ARP has limited impact on reconstruction metrics but degrades IDS and HS, demonstrating that these modules primarily improve reference utilization and entity-level fidelity. Comparing Rows 3, 4, and 5, removing WKS or WKF leads to consistent degradation in both full-reference and non-reference metrics, especially in IDS and HS. This verifies that the knowledge-related modules effectively enhance reference utilization and reduce hallucinations.
\begin{table}[h]
\centering
\caption{Ablation study on each world knowledge module}
\label{tab:ablation_world_knowledge}
\resizebox{\linewidth}{!}{%
\begin{tabular}{clcccccccc}
\toprule
& \textbf{Setting} & \textbf{PSNR}$\uparrow$ & \textbf{SSIM}$\uparrow$ & \textbf{LPIPS}$\downarrow$ & \textbf{RALI}$\uparrow$ & \textbf{Q-Insight}$\uparrow$ & \textbf{IDS}$\uparrow$ & \textbf{HS}$\uparrow$ \\
\midrule
\textcolor{gray}{\tiny 1} & w/o Iterative Refinement Loop (IRL) & 22.57 & 0.6727 & 0.3456 & 4.2636 & 4.1754 & 0.8080 & 4.30 \\
\textcolor{gray}{\tiny 2} & w/o Agentic ROI Planner (ARP) & 22.61 & 0.6722 & 0.3451 & 4.2593 & 4.1748 & 0.8033 & 4.09 \\
\textcolor{gray}{\tiny 3} & w/o World Knowledge Searcher (WKS) & 22.99 & 0.6798 & 0.3799 & 4.1397 & 4.0704 & 0.6894 & -- \\
\textcolor{gray}{\tiny 4} & w/o World Knowledge Filter (WKF) & 22.48 & 0.6636 & 0.3854 & 4.1280 & 4.0614 & 0.6577 & 3.94 \\
\textcolor{gray}{\tiny 5} & Full Module & \textbf{22.97} & \textbf{0.6879} & \textbf{0.3408} & \textbf{4.2522} & \textbf{4.1917} & \textbf{0.8199} & \textbf{4.47} \\
\bottomrule
\end{tabular}%
}
\end{table}

\paragraph{Effectiveness of World Knowledge Improvements} To disentangle the contribution of the Flux backbone from the proposed World Knowledge mechanism, we provide additional ablation experiments on the Face dataset, including Ada-RefSR, Ada-RefSR + World Knowledge, and our World-Grounded Image Super-Resolution Model (WGISR) with/without World Knowledge in \cref{tab:ablation_world_knowledge_improvements}. Comparing Rows 1 and 3, our model without reference images already achieves better non-reference metrics than Ada-RefSR, while Ada-RefSR obtains higher IDS due to its reference-based design. This shows that the improvement is not simply from the larger Flux backbone. More importantly, comparing Row 1 vs. Row 2 and Row 3 vs. Row 4, introducing World Knowledge consistently improves reconstruction quality, identity similarity (IDS), and hallucination score (HS). These results demonstrate that the performance gains come from effectively leveraging external knowledge rather than merely using a larger generative model.
\begin{table}[h]
\centering
\caption{Ablation study on world knowledge improvements.}
\label{tab:ablation_world_knowledge_improvements}
\resizebox{\linewidth}{!}{%
\begin{tabular}{clcccccccc}
\toprule
& \textbf{Setting} & \textbf{PSNR}$\uparrow$ & \textbf{SSIM}$\uparrow$ & \textbf{LPIPS}$\downarrow$ & \textbf{RALI}$\uparrow$ & \textbf{Q-Insight}$\uparrow$ & \textbf{IDS}$\uparrow$ & \textbf{HS}$\uparrow$ \\
\midrule
\textcolor{gray}{\tiny 1} & Ada-RefSR & \textbf{24.43} & \textbf{0.6998} & 0.3191 & 4.0267 & 3.8969 & 0.7364 & 3.85 \\
\textcolor{gray}{\tiny 2} & Ada-RefSR + World Knowledge & 24.42 & 0.6996 & \textbf{0.3171} & 4.0311 & 3.8973 & 0.7395 & 4.14 \\
\textcolor{gray}{\tiny 3} & WGISR w/o World Knowledge & 22.99 & 0.6798 & 0.3799 & 4.1397 & 4.0704 & 0.6894 & -- \\
\textcolor{gray}{\tiny 4} & WGISR + World Knowledge (Ours) & 22.97 & 0.6879 & 0.3408 & \textbf{4.2522} & \textbf{4.1917} & \textbf{0.8199} & \textbf{4.47} \\
\bottomrule
\end{tabular}%
}
\end{table}

\paragraph{Robustness of Mismatched Reference Images} We evaluate the robustness of WorldSR under mismatch references on the Face dataset by replacing the reference images with: (1) no reference, (2) irrelevant references (Bird images), (3) pure black images, (4) wrong face identities, and (5) noisy but correct references. The results are shown in \cref{tab:ablation_robustness}. For irrelevant references and pure black images (Rows 2–3), the performance remains close to the non-reference setting (Row 1), indicating that the model can distinguish unrelated references and avoid being significantly affected. For wrong identity references (Row 4), the model may inherit incorrect facial details from the reference, leading to degraded IDS and HS, which is expected for reference-guided SR. For weak references with noise (Rows 5–6), the model still extracts useful information from the references, achieving clear improvements in IDS and HS compared with the non-reference setting, with performance increasing as the reference quality improves. These results demonstrate that WorldSR is robust to irrelevant and degraded references, while effectively leveraging informative reference images to improve restoration.

\begin{table}[h]
\centering
\small
\caption{Robustness of mismatch reference images.}
\label{tab:ablation_robustness}
{%
\begin{tabular}{clccccc}
\toprule
& \textbf{Reference data} & \textbf{LPIPS}$\downarrow$ & \textbf{RALI}$\uparrow$ & \textbf{Q-Insight}$\uparrow$ & \textbf{IDS}$\uparrow$ & \textbf{HS}$\uparrow$ \\
\midrule
\textcolor{gray}{\tiny 1} & w/o Reference & 0.3799 & 4.1397 & 4.0704 & 0.6894 & -- \\
\textcolor{gray}{\tiny 2} & Irrelevant (bird image) & 0.3795 & 4.1472 & 4.1107 & 0.6808 & -- \\
\textcolor{gray}{\tiny 3} & Pure black image & 0.3650 & 4.1255 & 4.0912 & 0.6883 & -- \\
\textcolor{gray}{\tiny 4} & Wrong face identity & 0.3736 & 4.1895 & 4.1364 & 0.6347 & 2.51 \\
\textcolor{gray}{\tiny 5} & Correct reference, 10~dB noise & 0.3716 & 4.1659 & 4.1327 & 0.7578 & 4.16 \\
\textcolor{gray}{\tiny 6} & Correct reference, 20~dB noise & 0.3600 & 4.1987 & 4.1593 & 0.7794 & 4.35 \\
\textcolor{gray}{\tiny 7} & Correct reference & \textbf{0.3408} & \textbf{4.2522} & \textbf{4.1917} & \textbf{0.8199} & \textbf{4.47} \\
\bottomrule
\end{tabular}%
}
\end{table}

\subsection{More Qualitative Results}\label{subsec:more_visual}
\begin{figure}[H]
    \centering
    \includegraphics[width=1\textwidth]{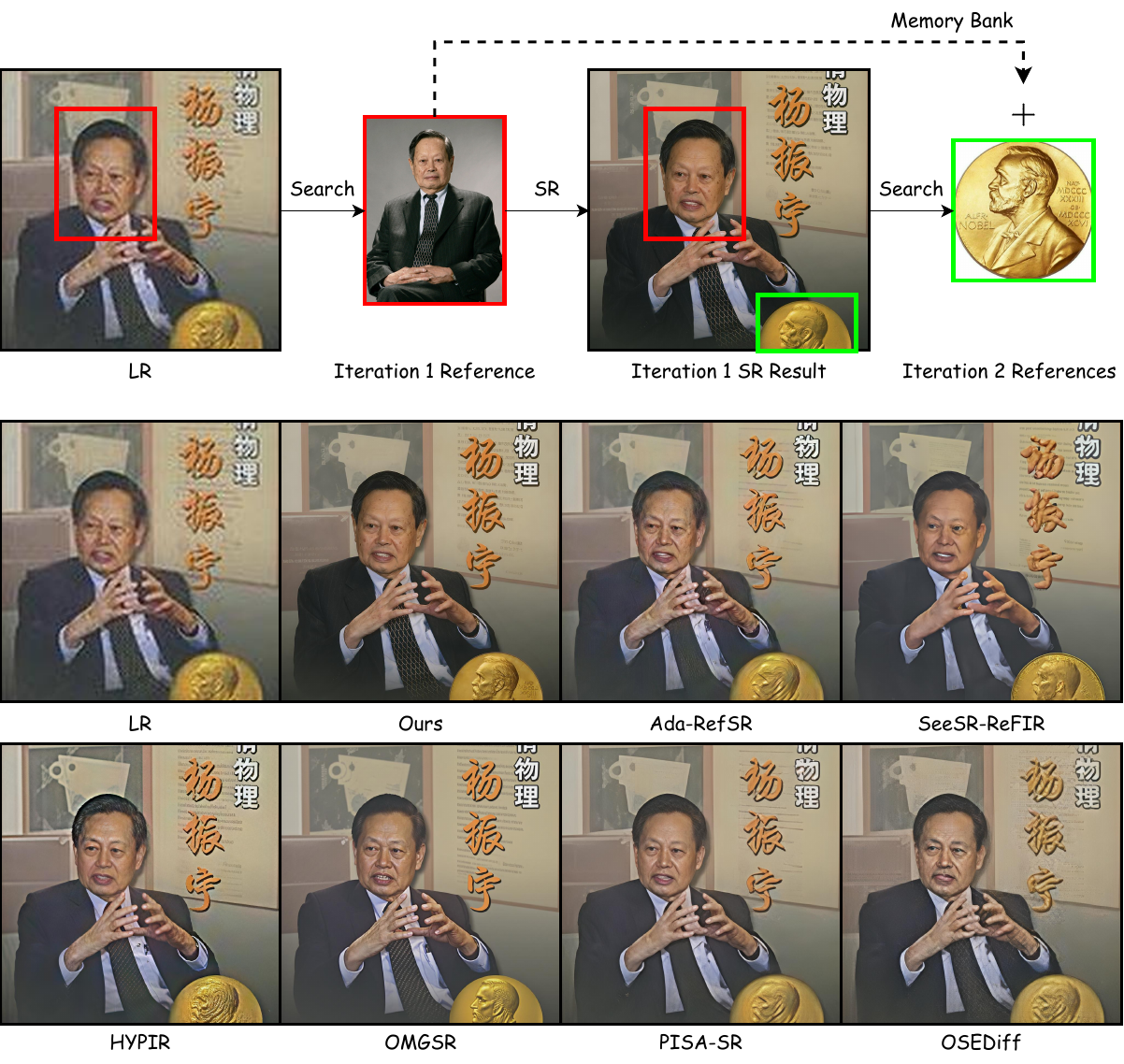}
    \caption{Qualitative comparison on the RealLQ250 dataset. Please zoom in for more details.}
    \label{fig:more_visual_RealLQ250_1}
\end{figure}

\begin{figure}[H]
    \centering
    \includegraphics[width=0.85\textwidth]{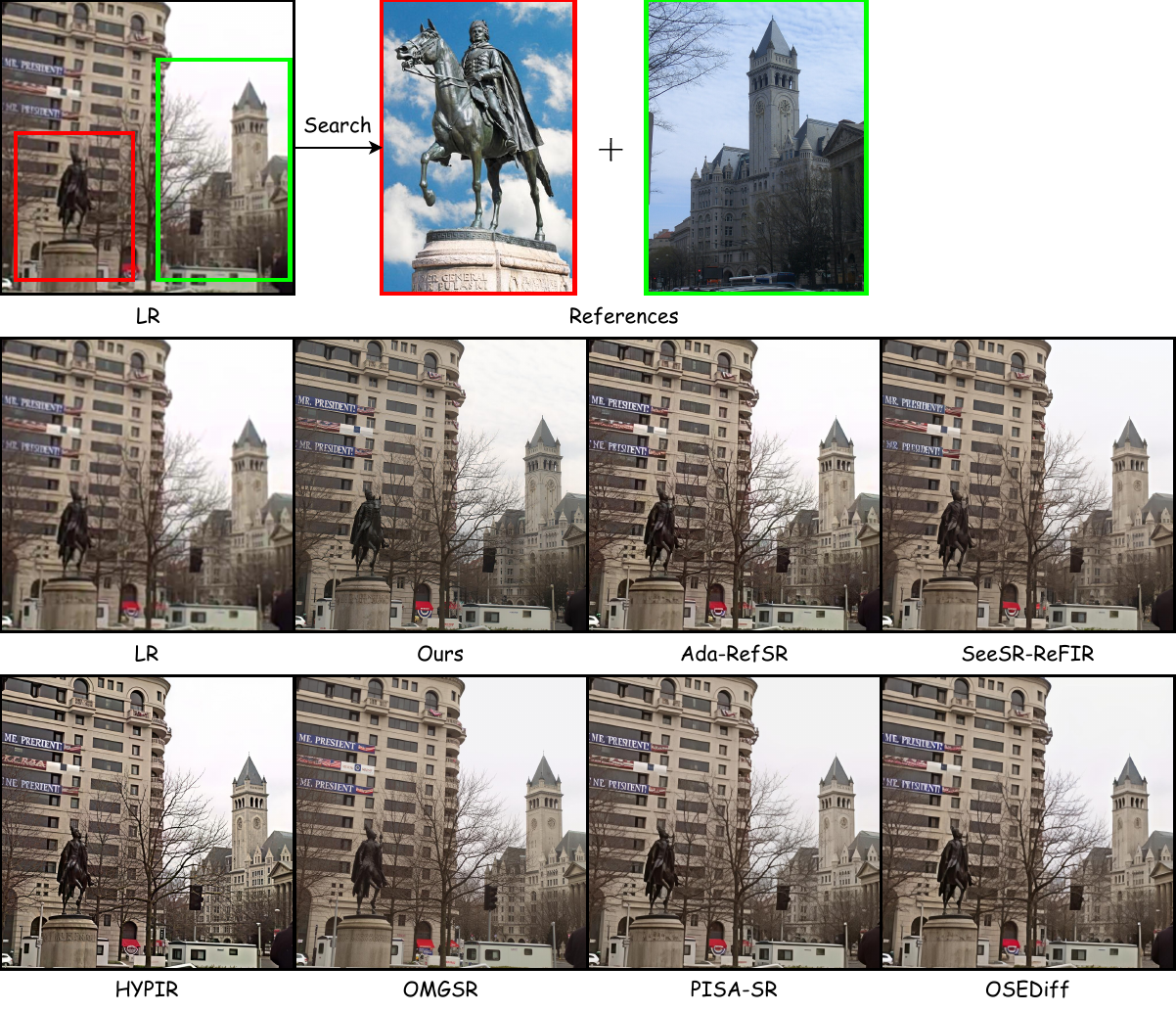}
    \caption{Qualitative comparison on the RealLQ250 dataset. Please zoom in for more details.}
    \label{fig:more_visual_RealLQ250_2}
\end{figure}

In this section, we provide additional qualitative results evaluated across the RealLQ250, Face, Bird, WRSR, and CUFED5 datasets. For the RealLQ250 dataset, we first illustrate the entity bounding boxes and search results, followed by qualitative comparisons with other methods, as shown in \cref{fig:more_visual_RealLQ250_1} and \cref{fig:more_visual_RealLQ250_2}. Specifically, \cref{fig:more_visual_RealLQ250_1} presents cases requiring multiple searches through our iterative refinement loop, while \cref{fig:more_visual_RealLQ250_2} demonstrates scenarios where reference knowledge for all entities is obtained in just a single loop. As the Face, Bird, WRSR, and CUFED5 datasets consist mostly of single entities, we provide only a selected reference image and visual comparisons for these datasets.
Ultimately, these visual comparisons demonstrate that our method significantly outperforms current SOTA methods in terms of perceptual quality and detail recovery. 

\begin{figure}[H]
    \centering
    \includegraphics[width=0.85\textwidth]{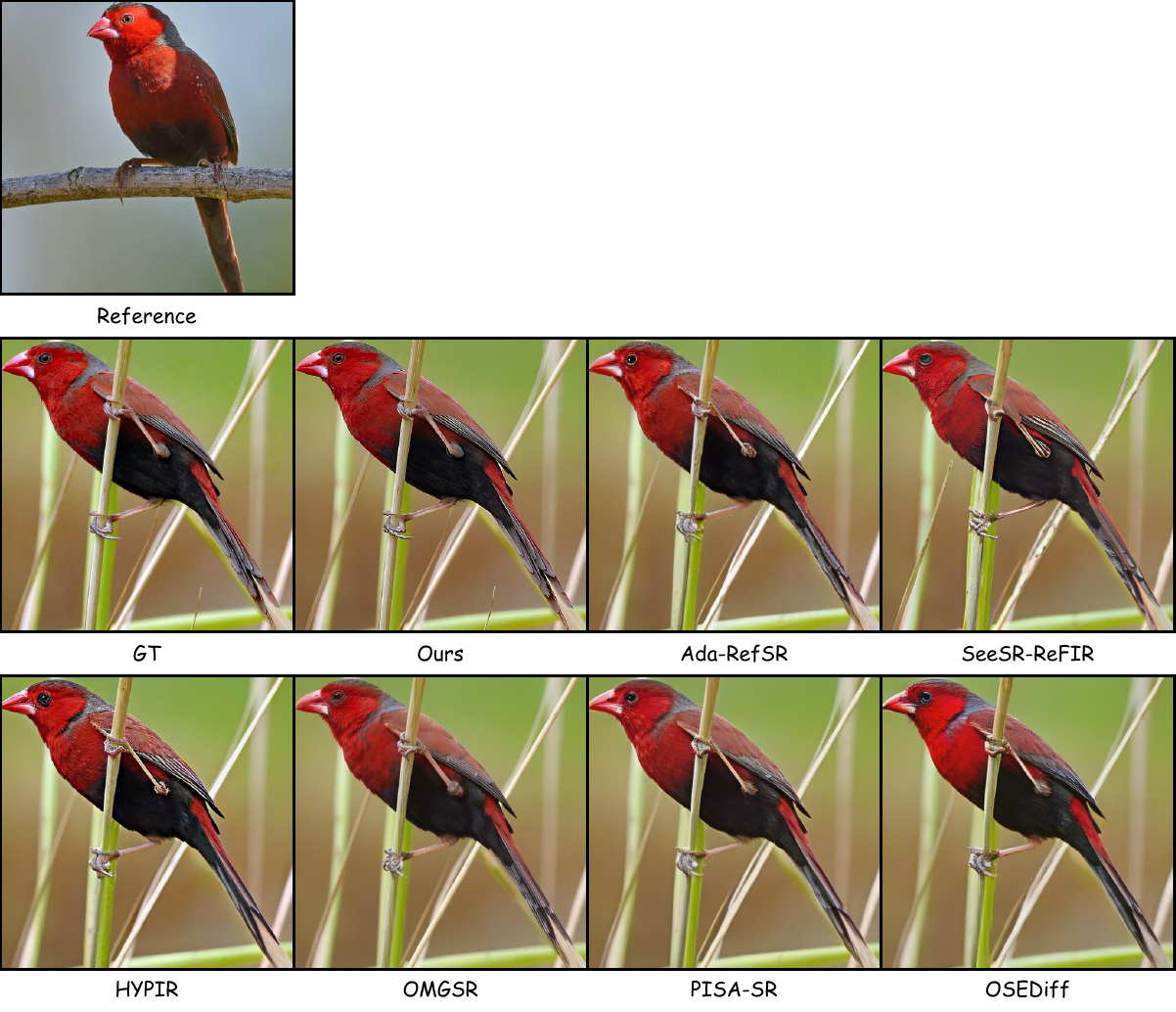}
    \caption{Qualitative comparison on the Bird dataset. Please zoom in for more details.}
    \label{fig:more_visual_Bird_1}
\end{figure}
\begin{figure}[H]
    \centering
    \includegraphics[width=0.85\textwidth]{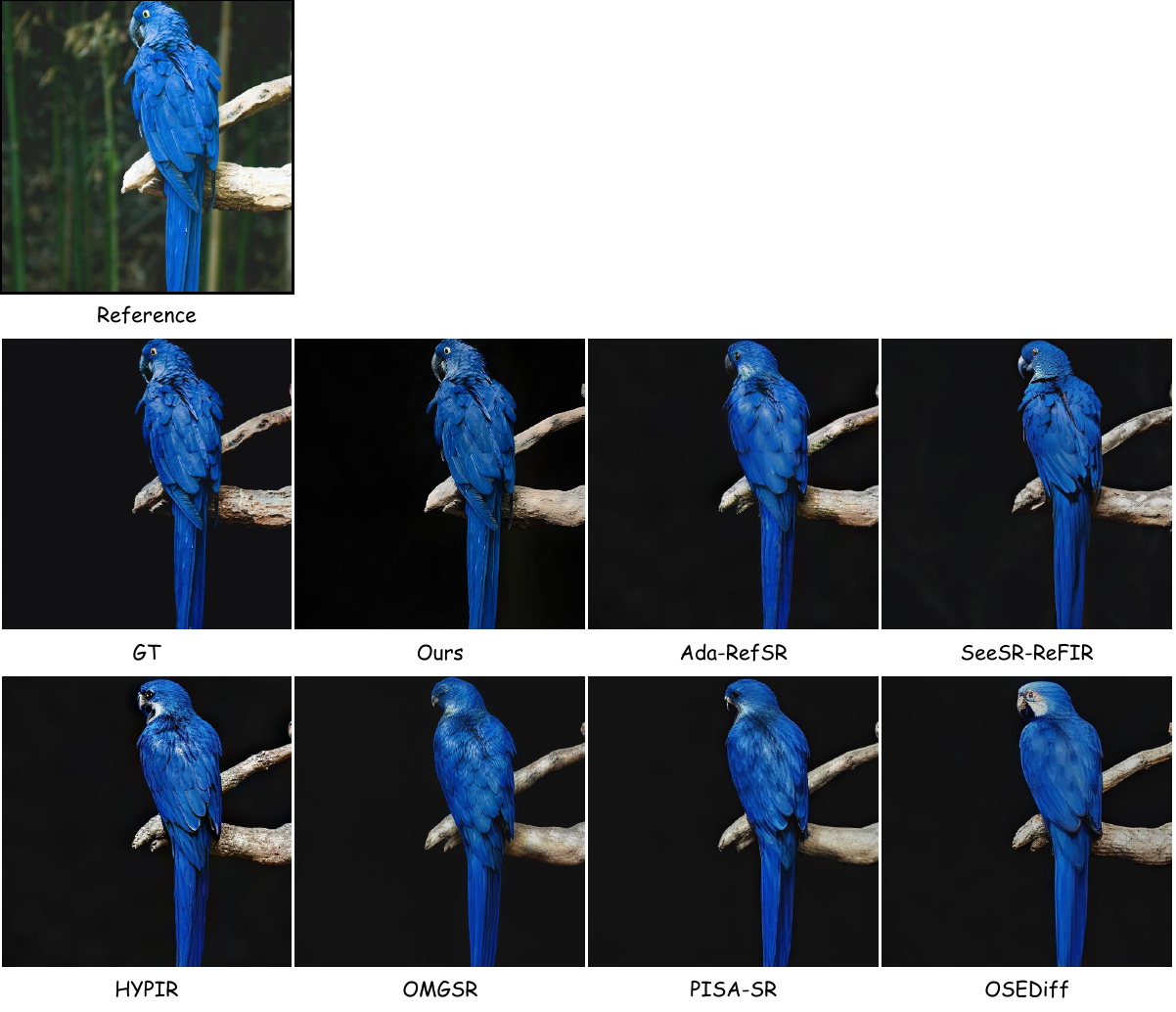}
    \caption{Qualitative comparison on the Bird dataset. Please zoom in for more details.}
    \label{fig:more_visual_Bird_2}
\end{figure}
\begin{figure}[H]
    \centering
    \includegraphics[width=0.85\textwidth]{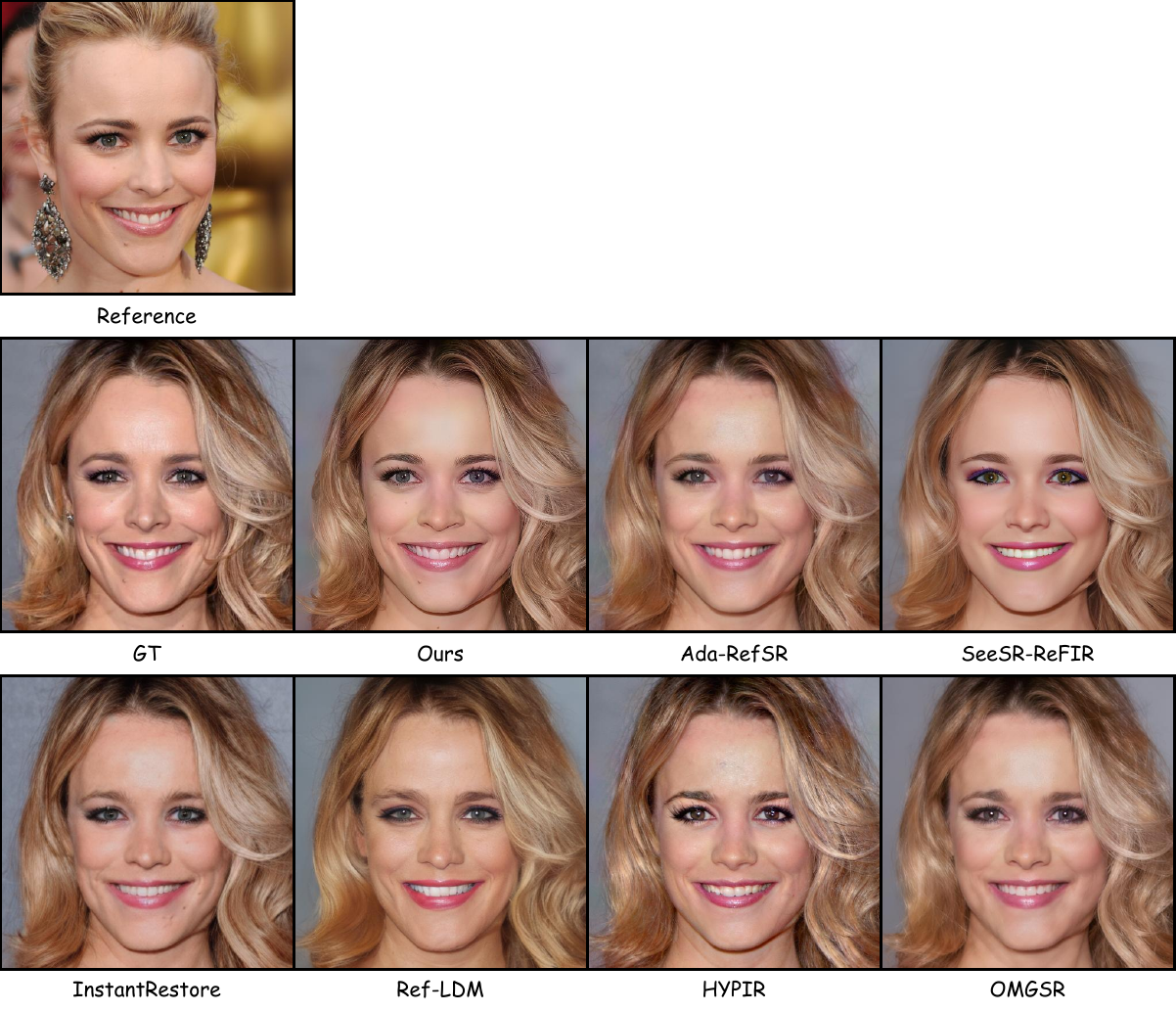}
    \caption{Qualitative comparison on the Face dataset. Please zoom in for more details.}
    \label{fig:more_visual_Face_1}
\end{figure}
\begin{figure}[H]
    \centering
    \includegraphics[width=0.85\textwidth]{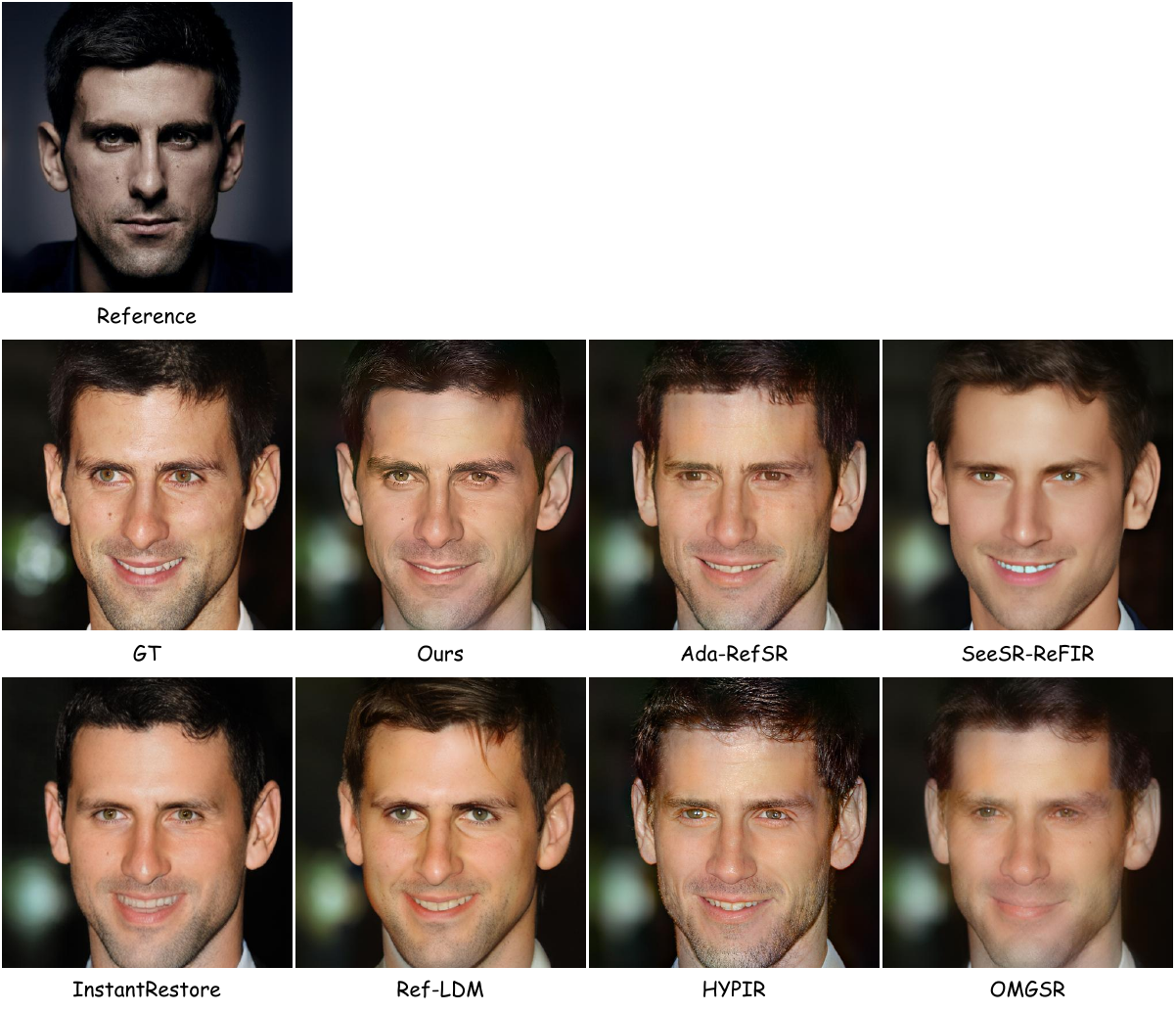}
    \caption{Qualitative comparison on the Face dataset. Please zoom in for more details.}
    \label{fig:more_visual_Face_2}
\end{figure}
\begin{figure}[H]
    \centering
    \includegraphics[width=0.85\textwidth]{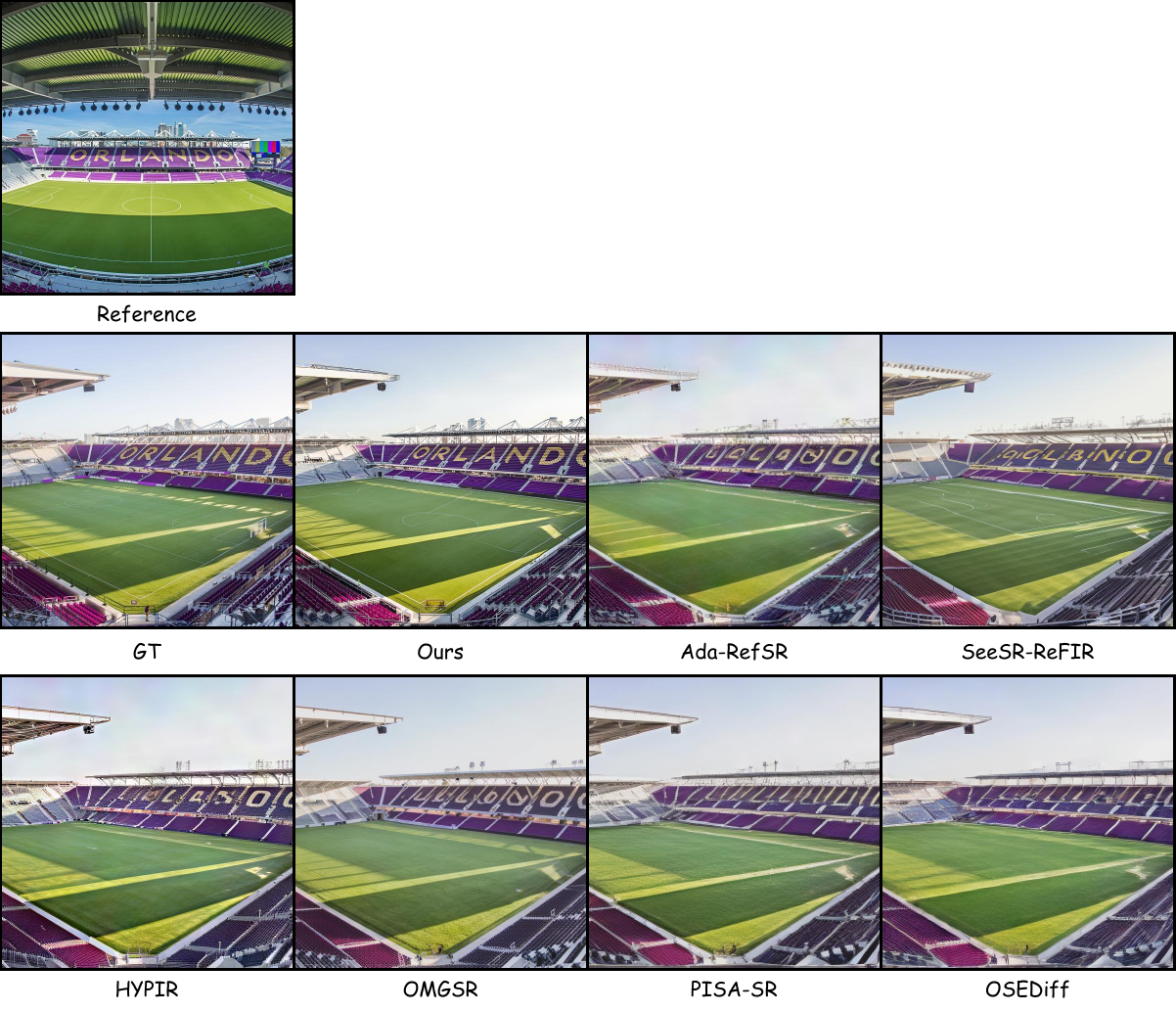}
    \caption{Qualitative comparison on the WRSR dataset. Please zoom in for more details.}
    \label{fig:more_visual_WRSR_1}
\end{figure}
\begin{figure}[H]
    \centering
    \includegraphics[width=0.85\textwidth]{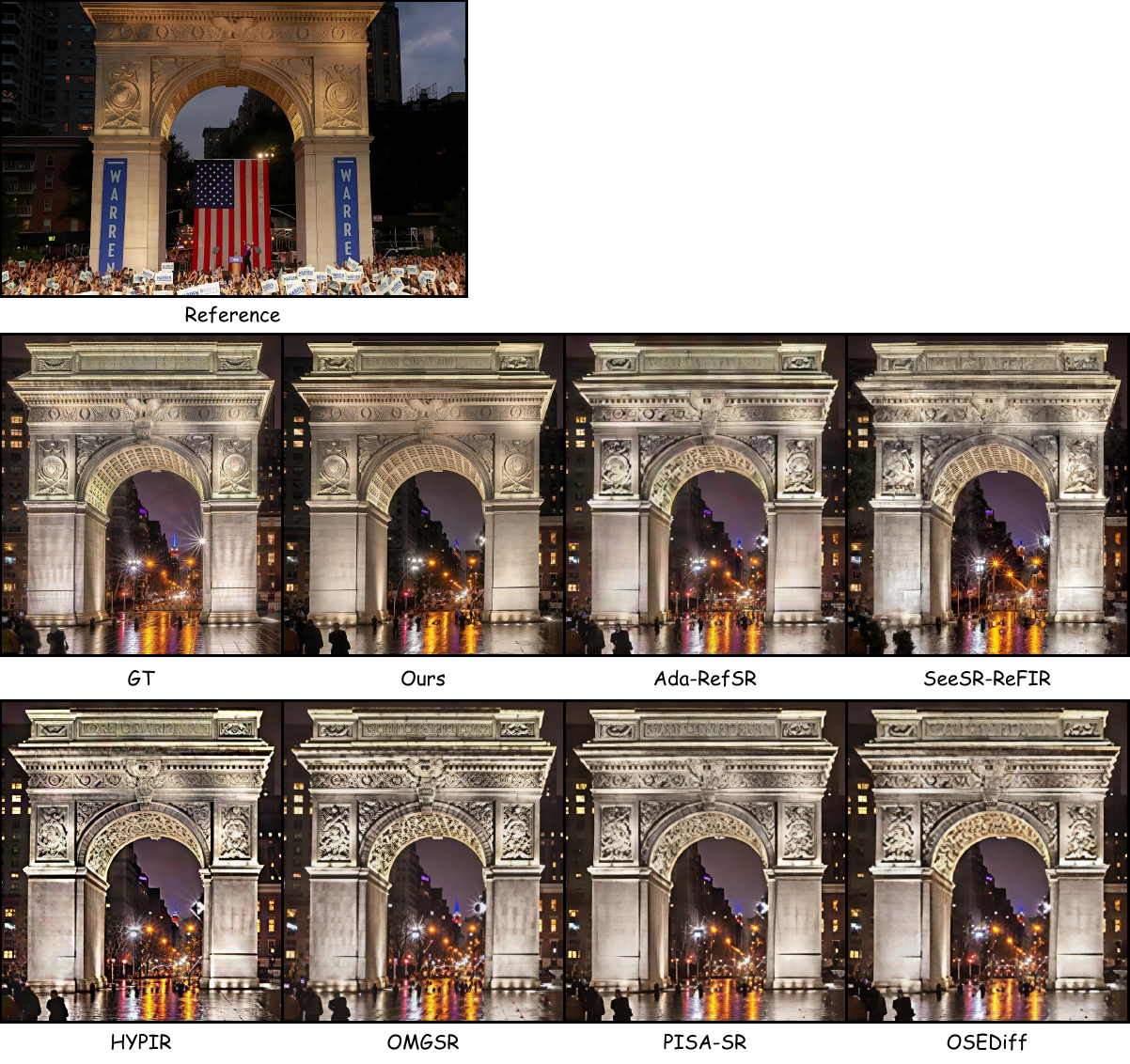}
    \caption{Qualitative comparison on the WRSR dataset. Please zoom in for more details.}
    \label{fig:more_visual_WRSR_2}
\end{figure}
\begin{figure}[H]
    \centering
    \includegraphics[width=0.85\textwidth]{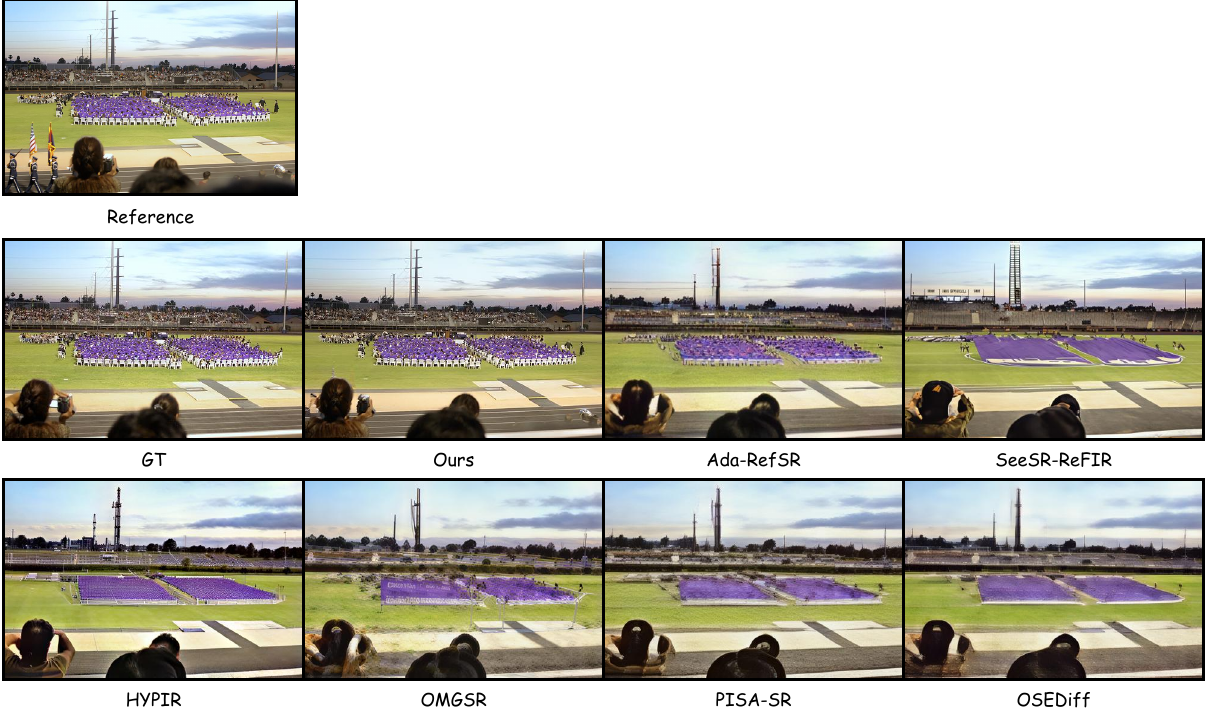}
    \caption{Qualitative comparison on the CUFED5 dataset. Please zoom in for more details.}
    \label{fig:more_visual_CUFED5_1}
\end{figure}
\begin{figure}[H]
    \centering
    \includegraphics[width=1.0\textwidth]{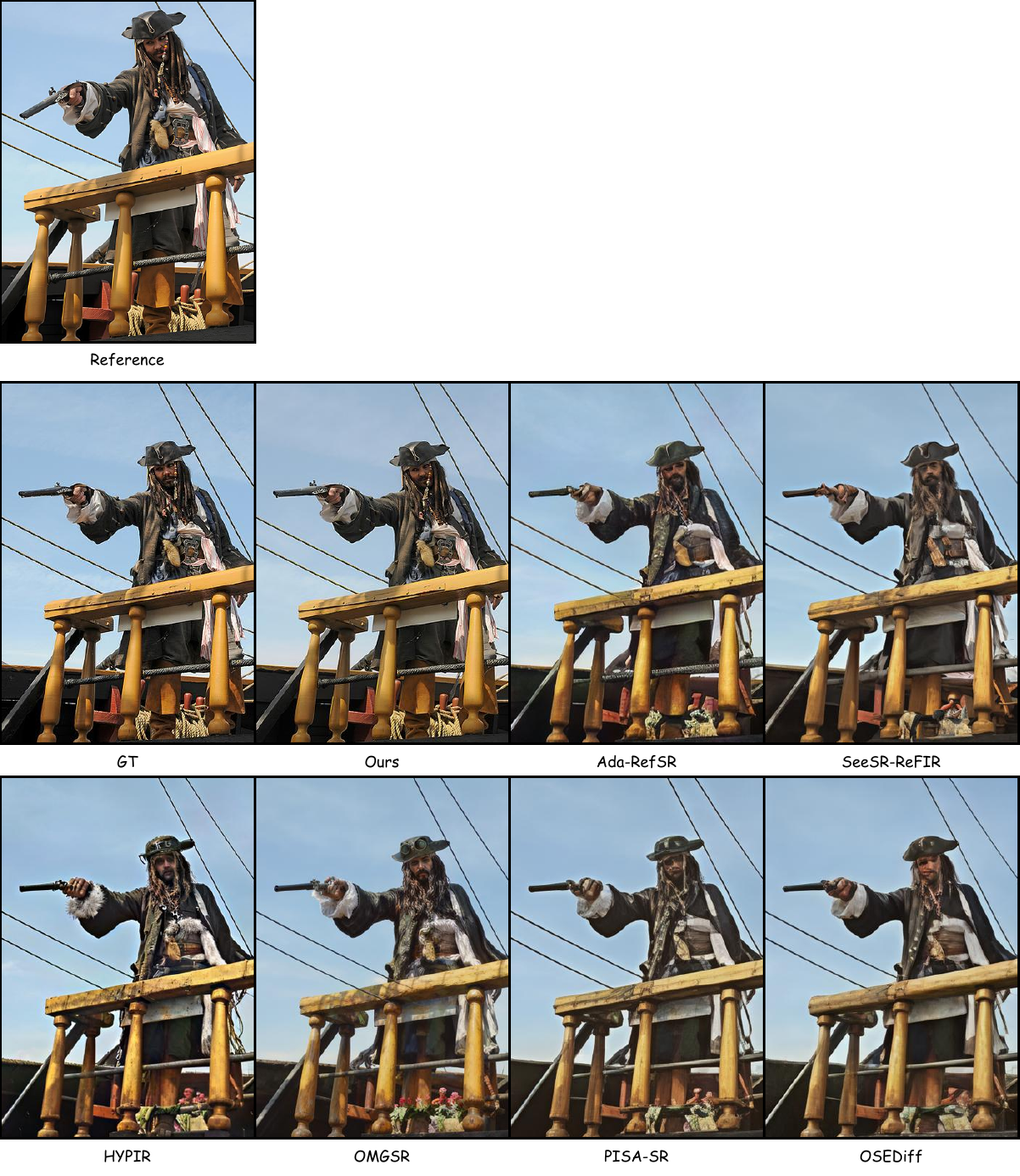}
    \caption{Qualitative comparison on the CUFED5 dataset. Please zoom in for more details.}
    \label{fig:more_visual_CUFED5_2}
\end{figure}



\newpage
\section*{NeurIPS Paper Checklist}

\begin{enumerate}

\item {\bf Claims}
    \item[] Question: Do the main claims made in the abstract and introduction accurately reflect the paper's contributions and scope?
    \item[] Answer: \answerYes{} 
    \item[] Justification: The abstract and introduction accurately reflect our paper's contributions and scope.

\item {\bf Limitations}
    \item[] Question: Does the paper discuss the limitations of the work performed by the authors?
    \item[] Answer: \answerYes{} 
    \item[] Justification: The limitations have been included in conclusion section.

\item {\bf Theory assumptions and proofs}
    \item[] Question: For each theoretical result, does the paper provide the full set of assumptions and a complete (and correct) proof?
    \item[] Answer: \answerNA{} 
    \item[] Justification: Our paper does not include theoretical results.

    \item {\bf Experimental result reproducibility}
    \item[] Question: Does the paper fully disclose all the information needed to reproduce the main experimental results of the paper to the extent that it affects the main claims and/or conclusions of the paper (regardless of whether the code and data are provided or not)?
    \item[] Answer: \answerYes{} 
    \item[] Justification: We have include the implementation details in the Experiments Section and the Supplemental materials.

\item {\bf Open access to data and code}
    \item[] Question: Does the paper provide open access to the data and code, with sufficient instructions to faithfully reproduce the main experimental results, as described in supplemental material?
    \item[] Answer: \answerNo{} 
    \item[] Justification: The source code is not currently open-sourced due to policy restrictions.

\item {\bf Experimental setting/details}
    \item[] Question: Does the paper specify all the training and test details (e.g., data splits, hyperparameters, how they were chosen, type of optimizer) necessary to understand the results?
    \item[] Answer: \answerYes{} 
    \item[] Justification: The training and test details can be found in the Experiments Section and the Supplemental materials.

\item {\bf Experiment statistical significance}
    \item[] Question: Does the paper report error bars suitably and correctly defined or other appropriate information about the statistical significance of the experiments?
    \item[] Answer: \answerNo{}  
    \item[] Justification: The error bars are not reported because it would be too computationally expensive.

\item {\bf Experiments compute resources}
    \item[] Question: For each experiment, does the paper provide sufficient information on the computer resources (type of compute workers, memory, time of execution) needed to reproduce the experiments?
    \item[] Answer: \answerYes{} 
    \item[] Justification: The information of used computer resources can be found in the Experiments Section and the Supplemental materials.
    
\item {\bf Code of ethics}
    \item[] Question: Does the research conducted in the paper conform, in every respect, with the NeurIPS Code of Ethics \url{https://neurips.cc/public/EthicsGuidelines}?
    \item[] Answer: \answerYes{}  
    \item[] Justification: This work is conducted in accordance with the NeurIPS Code of Ethics.

\item {\bf Broader impacts}
    \item[] Question: Does the paper discuss both potential positive societal impacts and negative societal impacts of the work performed?
    \item[] Answer: \answerNA{} 
    \item[] Justification: There is no societal impact of the work performed.
    
\item {\bf Safeguards}
    \item[] Question: Does the paper describe safeguards that have been put in place for responsible release of data or models that have a high risk for misuse (e.g., pre-trained language models, image generators, or scraped datasets)?
    \item[] Answer: \answerNA{} 
    \item[] Justification:  The paper poses no any risks.

\item {\bf Licenses for existing assets}
    \item[] Question: Are the creators or original owners of assets (e.g., code, data, models), used in the paper, properly credited and are the license and terms of use explicitly mentioned and properly respected?
    \item[] Answer: \answerYes{} 
    \item[] Justification: We properly credited all related works.

\item {\bf New assets}
    \item[] Question: Are new assets introduced in the paper well documented and is the documentation provided alongside the assets?
    \item[] Answer: \answerNA{} 
    \item[] Justification: The paper does not release new assets.

\item {\bf Crowdsourcing and research with human subjects}
    \item[] Question: For crowdsourcing experiments and research with human subjects, does the paper include the full text of instructions given to participants and screenshots, if applicable, as well as details about compensation (if any)? 
    \item[] Answer: \answerNA{}  
    \item[] Justification: The paper does not involve crowdsourcing nor research with human subjects.

\item {\bf Institutional review board (IRB) approvals or equivalent for research with human subjects}
    \item[] Question: Does the paper describe potential risks incurred by study participants, whether such risks were disclosed to the subjects, and whether Institutional Review Board (IRB) approvals (or an equivalent approval/review based on the requirements of your country or institution) were obtained?
    \item[] Answer: \answerNA{} 
    \item[] Justification: The paper does not involve crowdsourcing nor research with human subjects.

\item {\bf Declaration of LLM usage}
    \item[] Question: Does the paper describe the usage of LLMs if it is an important, original, or non-standard component of the core methods in this research? Note that if the LLM is used only for writing, editing, or formatting purposes and does \emph{not} impact the core methodology, scientific rigor, or originality of the research, declaration is not required.
    \item[] Answer: \answerNA{}  
    \item[] Justification: The core method development in this research does not involve LLMs as any important, original, or non-standard components.

\end{enumerate}

\end{document}